\documentclass[journal]{IEEEtran}
\usepackage[utf8]{inputenc}
\usepackage{amsmath,amssymb,amsfonts}
\usepackage[nospace]{cite}

\usepackage{algorithmic}
\usepackage{multirow}

\usepackage{graphicx}
\usepackage{textcomp}
\usepackage{xcolor}
\usepackage{pifont}
\usepackage{amsthm}
\usepackage{hyperref}
\usepackage{mathrsfs}
\usepackage{booktabs}
\usepackage{hyperref}
\usepackage{cases}
\usepackage{comment}
\usepackage{empheq} 
\usepackage{caption}
\usepackage{subcaption}
\usepackage{fancyhdr}

\allowdisplaybreaks

\usepackage{url}  
\usepackage{graphicx}  
\usepackage[ruled,vlined,linesnumbered]{algorithm2e}
\usepackage{setspace}
\usepackage{floatpag} 
\usepackage{float}
\usepackage{bm}
\floatpagestyle{empty}

\usepackage{listings}

\title{TranDRL: A Transformer-Driven Deep Reinforcement Learning Enabled Prescriptive Maintenance Framework}

\author{
\IEEEauthorblockN{Yang Zhao,~\IEEEmembership{Member,~IEEE}, Jiaxi Yang,~\IEEEmembership{Student Member,~IEEE},
Wenbo~Wang,~\IEEEmembership{Senior Member,~IEEE},
Helin~Yang,~\IEEEmembership{Senior Member,~IEEE},
Dusit~Niyato,~\IEEEmembership{Fellow,~IEEE}
}

\thanks{This research is supported in part by National Natural Science Foundation of China under grant No. 62302046, in part by National Natural Science Foundation of China under Yunnan Major Scientific and Technological Projects under Grant 202202AG050002, in part by the National Key Research and Development Program of China under grant No. 2023YFB3308401, and in part by ChingMu Tech. Ltd. Research Project ``WiTracker’’ under grant No. KKF0202301252, in part by the National Research Foundation, Singapore, and Infocomm Media Development Authority under its Future Communications Research $\&$ Development Programme, Defence Science Organisation (DSO) National Laboratories under the AI Singapore Programme (AISG Award No: AISG2-RP-2020-019 and FCP-ASTAR-TG-2022-003), Singapore Ministry of Education (MOE) Tier 1 (RG87/22), and the NTU Centre for Computational Technologies in Finance (NTU-CCTF).}

\thanks{Yang Zhao is with Singapore Institute of Manufacturing Technology (SIMTech), Agency for Science, Technology and Research (A$*$STAR), 2 Fusionopolis Way, Innovis $\#$08-04, Singapore 138634, Republic of Singapore.
 (Email: {zhao\_yang@simtech.a-star.edu.sg).}
}
\thanks{Jiaxi Yang is with the Shenzhen Institute for Advanced Study \& School of Computer Science and Engineering, University of Electronic Science and Technology of China, Chengdu, China. (Email: {abbottyang@std.uestc.edu.cn}).}

\thanks{Wenbo Wang is with the Faculty of Mechanical and Electrical Engineering and Yunnan Key Laboratory of Intelligent Control and Application, Kunming University of Science and Technology, Kunming, China, 650050. (Email: \mbox{wenbo wang@kust.edu.cn}). (corresponding author)}

\thanks{Helin Yang is with the Department of Information and Communication Engineering, Xiamen University, Xiamen, China 361005 (Email: {helinyang066@xmu.edu.cn}.}

\thanks{Dusit Niyato is with the School of Computer Science and Engineering, Nanyang Technological University, Singapore, 639798. (Email: dniyato@ntu.edu.sg).}
}

\begin{document}     

\maketitle

\begin{abstract}
Industrial systems require reliable predictive maintenance strategies to enhance operational efficiency and reduce downtime. Existing studies rely on heuristic models which may struggle to capture complex temporal dependencies. This paper introduces an integrated framework that leverages the capabilities of the Transformer and Deep Reinforcement Learning (DRL) algorithms to optimize system maintenance actions. Our approach employs the Transformer model to effectively capture complex temporal patterns in IoT sensor data, thus accurately predicting the Remaining Useful Life (RUL) of equipment. Additionally, the DRL component of our framework provides cost-effective and timely maintenance recommendations. Numerous experiments conducted on the NASA C-MPASS dataset demonstrate that our approach has a performance similar to the ground-truth results and could be obviously better than the baseline methods in terms of RUL prediction accuracy as the time cycle increases. Additionally, experimental results demonstrate the effectiveness of optimizing maintenance actions.

\end{abstract}

\section{Introduction}
Industry 4.0 has introduced unprecedented levels of automation and data exchange in manufacturing technologies~\cite{doi:10.1080/00207543.2018.1444806}. However, with these advancements comes increased requirement for reliable, efficient, and cost-effective maintenance systems. Prescriptive maintenance has thus emerged as a crucial topic in both research and industrial sectors, offering the promise of proactive fault detection and optimized operation strategy planning~\cite{dangut2022application}. It differs from reactive maintenance, which only kicks in post-failure, causing often unplanned and expensive downtimes. According to ISO standard 13381-1~\cite{5413482}, the predictive maintenance cycle includes data preprocessing, failure prognostics, and action planning. However, current predictive maintenance systems largely rely on heuristic models for action planning or conventional machine learning algorithms for sensor data processing. Although effective to some extent, existing approaches often fail to capture complex temporal dependencies in sensor data due to issues such as vanishing or exploding gradients, limiting their ability to accurately predict RUL.
To address these issues, a number of new frameworks (e.g., attention in recurrent networks~\cite{aitken2021understanding}) are developed to capture the correlation in temporal sequence data.

Apart from that, there is a trend in recent studies to leverage DRL to make plans in maintenance tasks~\cite{ong2021deep}. DRL excels at making optimal decisions in complex environments, offering a powerful tool for action planning in maintenance tasks. As a result, the DRL-based decisions offer optimal maintenance action schedules that balance the trade-off between equipment reliability and maintenance costs. However, these methods consistently encounter the cold start problem~\cite{NIPS2017_faafda66}, coupled with low learning efficiency. In some special cases, such as aero-engines, these methods also suffer from a lack of reliability and explainability. This often leads to their decisions being perceived as untrustworthy by users.

Unlike existing solutions, we propose a framework named \textsc{TranDRL}, which integrates both aspects seamlessly. \textsc{TranDRL}  leverages both the Transformer framework and DRL to not only accurately predict the RUL but also convert these predictions into maintenance action recommendations. Additionally, we adopt the RL with Human Feedback (RLHF) algorithm, which incorporates the human expert's opinion on maintenance actions into the DRL algorithm to facilitate the algorithm in making decisions. By integrating these two critical neural networks and human expert's experience, our framework provides a comprehensive solution for industrial maintenance challenges. Transformers excel at modeling long-term dependencies in sequential data through their attention mechanism and exhibit a high degree of adaptability, making them ideal for handling complex industrial sensor data. Utilizing the expertise and insights of human maintenance professionals, the algorithm can efficiently guide the production operational actions, expediting the convergence of the DRL algorithm toward optimal maintenance strategies. This human-guided approach ensures that the algorithm remains on the right track, continually improving its performance and leading to more efficient and cost-effective industrial maintenance practices.

\textbf{Contributions.} The main contributions of this paper can be summarized as follows:
\begin{itemize}
    \item To improve the accuracy of the RUL prediction module, we adopt a Transformer-based neural network to address the difficulty in reflecting long-term dependencies of sequential sensor data.
    \item We propose a DRL-based recommendation approach for determining maintenance actions, reducing both labor involvement and operational costs. Furthermore, we utilize the RLHF algorithm to confirm the correct maintenance actions and improve the learning efficiency of the DRL framework.
   \item The transformer-enabled DRL algorithm is tailored into a federated learning framework, which can optimize the prescriptive maintenance process in multi-machine scenarios.
    \item Extensive simulation experiment results demonstrate that our method outperforms a series of baseline methods.
\end{itemize}

\textbf{Organizations.} The rest of the paper is organized as follows. Section~\ref{related-work} reviews related work in the field of prescriptive maintenance utilizing machine learning algorithms. Section~\ref{problem} formulates the problem we are trying to solve. Then, Section~\ref{framework}  describes our proposed framework for Transformer-driven prediction maintenance in detail. Next, Section~\ref{sec:case} presents a case study on our proposed framework using a real-world dataset. Finally, Section~\ref{conclusion} concludes the paper and outlines future directions for research. Notations used across the paper are provided in Table~\ref{notations}.

\begin{table}[!ht]
\centering
\caption{Notations and abbreviations.}
\label{notations}
\begin{tabular}{cp{6.5cm}}
\toprule[1pt]
\textbf{Symbol} & \textbf{Descriptions} \\
\midrule
$\mathbf{X}$ & Time series of sensor readings and operational parameters. \\
$R(t)$ & Remaining Useful Life at time $t$: $R(t) = T - t$. \\
$T$ & Total expected lifespan of the component/machine. \\
$t$ & Current operational time of the machine in concern. \\
$M$ & Model estimating $R(t)$ from $\mathbf{X}$. \\
$s$ & State encapsulating the health and RUL metrics. \\
$\mathcal{A}$ & Set of maintenance actions. \\
$a$ & Action chosen to maximize the future reward. \\
$a^*$ & Optimal action maximizing the cumulative benefit. \\
$Q(s, a)$ & Expected reward of action $a$ in state $s$. \\
$\epsilon$ & Exploration rate in DRL. \\
$\alpha$ & Learning rate in DRL. \\
$\gamma$ & Discount factor for future rewards. \\
$\boldsymbol{y}$ & Predicted RUL output by the neural network model. \\
$\boldsymbol{s}$ & Observation state in DRL from RUL predictions. \\
$d_{\text{model}}$ & Input/output dimension of the Transformer. \\
$d_k, d_v$ & Dimensions of keys/queries and values in attention computation. \\
$\mathbf{Q}, \mathbf{K}, \mathbf{V}$ & Query, key, and value matrices in attention computation. \\
$FFN(\cdot)$ & Feed-forward network in the Transformer module. \\
$\bm{\Lambda}$ & Attention weight matrix. \\
$\alpha_{ij}$ & Attention weight from the $j$-th input to the $i$-th output. \\
$\boldsymbol{h}$ & Hidden states in the encoder. \\
$\textbf{O}$ & Output states from the decoder. \\
$\sigma(\cdot)$ & Activation function in the prediction module. \\
$K$ & Number of machines in federated learning setups. \\
$f_{\text{ensemble}}$ & Function for RUL prediction in the multi-machine case. \\
$\mathbf{W}, b$ & Weight matrix and bias in the prediction module. \\
PCA & Principal component analysis. \\
DQN & Deep Q-network. \\
PPO & Proximal Policy Optimization. \\
SAC & Soft Actor-Critic. \\
\bottomrule[1pt]
\end{tabular}
\end{table}
\section{Related Work}
\label{related-work}
Prescriptive maintenance uses historical operational data and (possibly implicit) system models to optimize maintenance timing and strategies. It extends the prediction maintenance approaches by optimizing strategies based on fault diagnosis, focusing on identifying root causes and selecting proper maintenance actions~\cite{zonta2020predictive}. Compared to traditional methods, it aims to prevent waste and production losses by leveraging historical system status to forecast future machine status and optimize maintenance resources.

\subsection{Deep Learning-based Remaining Useful Life Prediction}

In production maintenance tasks, accurately predicting the RUL of machinery has traditionally required intricate models that capture the nuances of production system operations. While many revert to statistical signal processing and machine learning to bypass comprehensive system modeling, these methods often require deep domain-specific knowledge, especially for tasks such as signal denoising and feature extraction~\cite{zhou2014remaining, man2018prediction}. Notably, they also entail labor-intensive manual feature engineering. However, with machines becoming multifaceted and sensor-rich, these methods grapple with high-dimensional data. Deep learning, boosted by computational advances, emerges as a remedy. Pioneering algorithms, such as Convolutional Neural Networks (CNN) and Deep Autoencoders, have gained popularity owning to their proficiency in fault diagnosis and predictive maintenance~\cite{jia2016deep, zhao2019deep}. Their intrinsic capability to distill hierarchical features from raw data eliminates the need of manual feature engineering, thus fostering more accurate RUL predictions.

Amongst the deep learning architectures, Transformers, originally conceived for language processing tasks~\cite{vaswani2017attention}, have recently demonstrated efficacy in time-series data processing. For instance, Zhou \emph{et al.}~\cite{zhou2021informer} integrated a time-embedding layer, augmenting the Transformer's time series prediction. Yet, a full exploration of Transformers in the realm of prescriptive maintenance remains on the horizon.

\subsection{Deep Reinforcement Learning-enabled Predictive Maintenance}
DRL's concept of predictive maintenance has primarily focused on refining maintenance schedules~\cite{huang2020deep, dangut2022application, ong2021deep}. While these applications exhibit DRL's capability in mitigating machinery breakdowns, they predominantly center on forecasting malfunctions, focusing on actionable preventative measures. Recent work such as~\cite{gordon2022data} explored the possibility of employing DRL for prescriptive maintenance, but the integration of prediction and actionable recommendations remained elusive.

In existing maintenance studies, researchers have explored a number of preventive and predictive methods, as mentioned in~\cite{ansari2019prima} and~\cite{nemeth2018prima}. However, there is a growing interest in prescriptive maintenance, which extends beyond merely predicting problems, instead offering suggestions for the best actions to take. Studies such as~\cite{ansari2019prima} and~\cite{nemeth2018prima} have started to light the way, with most efforts focused on either predicting what will happen or prescribing what to do. However, even with improvements in using the Transformer technology for predictions and DRL for making recommendations, there is yet no plan to integrate these two ideas. Our study aims to fill this gap by introducing a way to utilize both the forecasting power of Transformers and the decision-making capabilities of DRL to not only predict equipment failures but also suggest the best actions to prevent them.

Furthermore, inspired by preference theory~\cite{hakim1998developing}, RLHF improves DRL performance by leveraging human feedback, reducing the difficulty in designing reward functions~\cite{christiano2017deep}. To address limitations of information and expensive data collection, expert demonstration and trajectory preferences are combined in~\cite{ibarz2018reward}, enabling direct human-agent interaction. Subsequent studies enhanced sample efficiency through semi-supervised learning and data augmentation~\cite{lee2021pebble, park2022surf}. In~\cite{early2022non} the non-Markov assumption is relaxed, formulating the scenario as a multiple-instance learning process. In safety-critical applications such as aero-engine maintenance, human feedback can set safe boundaries, therfore ensuring that the behavior of an agent remains within acceptable limits, making it more predictable and aligned with human expectations, as well as enhancing explainability and trust.

\section{Problem Formulation}
\label{problem}

In the modern industrial ecosystem, machinery and equipment form the backbone of operations. However, with increased automation and reliance on these machines, any unplanned downtime can be a critical blow to productivity and financial stability. The cascade effect is not just limited to production loss; it also affects aspects such as increased costs for emergency repairs, contractual penalties due to delivery delays, and even reputation damage~\cite{patti2008shape}. Therefore, the primary challenge is to recommend maintenance actions based on accurate machine failure prediction effectively.

\subsection{Remaining Useful Life Prediction}

Predicting RUL is essential for predictive maintenance, which provides an estimation of the duration that a machine or its components will continue to function before encountering a failure. An accurate RUL prediction enables industries to strategically schedule their maintenance activities. The RUL offers a dynamic and predictive window into the operational health of a machine. The operational data of a machinery component is typically represented as a time series $\mathbf{X} = [\boldsymbol{x_1}, \boldsymbol{x_2}, \ldots, \boldsymbol{x_n}]$, where each $\boldsymbol{x}_i$ can represent multiple sensor readings, operational parameters, or environmental conditions at a sampling instance $i$. The RUL at any given time $t$, denoted by $R(t)$, can be mathematically represented as:
$R(t) = T - t$,
where
\begin{itemize}
    \item $T$ is the total expected operational lifespan of the machine or component, which can vary based on its usage, environmental factors, and maintenance history.
    \item $t$ represents the machine's current age or operational time.
\end{itemize}
The inherent complexity is that $T$ is not a constant but a variable that might change based on multiple factors such as machine conditions. The primary challenge is to develop a predictive model $M$ that can use the time-series data $\mathbf{X}$ to provide an  approximation of $R(t)$:
$M(\mathbf{X}) \approx R(t)$.

\subsection{Optimal Maintenance Action Recommendation}

While predicting the RUL can help industries predict potential downtimes, it is the actionable insights derived from these predictions that empower industries to make informed decisions. Upon gaining prediction results of potential machine downtimes, the challenge shifts to determining the most optimal maintenance action. Leveraging the capabilities of DRL, our framework not only predicts the machinery's future status but also suggests the best course of action to extend its working time and ensure optimal performance.

Consider a machinery component's state $s$, which encapsulates its health, RUL, and other relevant metrics. From the state, we want to choose an action $a$ from a set of potential actions set $\mathcal{A}$ to maximize the expected future reward. 


Therefore, the challenge of this problem is to maximize delayed reinforcements. We both present and denote a value function $Q(s, a)$ that expresses the sum of the utility or profit from choosing $a$ in $s$. The objective is to navigate the action space and select:
$a^* = \arg\max_{a \in \mathcal{A}} Q(s, a)$,
where
\begin{itemize}
    \item $ Q(s, a)$ reflects immediate benefits and projected future rewards, considering the dynamics of machinery health and operation.
    \item The cumulative reward should integrate the current and the weighted future benefits.
\end{itemize}

\begin{figure*}[!ht]
    \centering
    {\includegraphics[width=\textwidth]{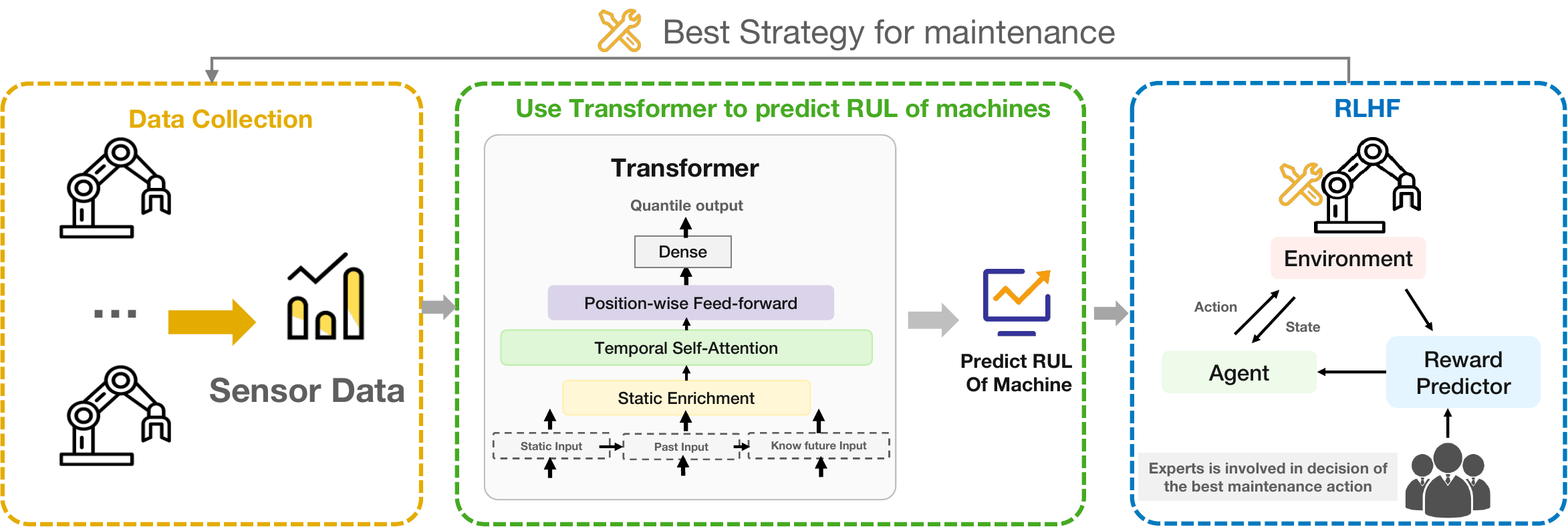}}
    \caption{An Overview of TranDRL Framework.}
    \label{fig:framework}
\end{figure*}

\section{Proposed Framework}
\label{framework}
To solve the problem above, we propose a framework consisting of two major components: a transformer-based learning model to predict the RUL and a decision module to recommend maintenance actions, as illustrated in Algorithm~\ref{algorithm-outline} and Figure~\ref{fig:framework}. Specifically, the sensor data is collected and utilized to predict the RUL of machines, after which DRL algorithms are employed to determine the optimal action for repairing machines.

\begin{algorithm}[!ht]
    \caption{TranDRL:A Transformer-enabled DRL framework for optimal maintenance action recommendation.}
    \label{algorithm-outline}
    \SetAlgoLined
    \KwIn{Sequence of sensor readings $\mathbf{X} = (\boldsymbol{x_1}, \boldsymbol{x_2}, \ldots, \boldsymbol{x_n})$, exploration rate $\epsilon$, learning rate $\alpha$, discount factor $\gamma$}
    \KwOut{Recommended maintenance action $a$}

    \tcc{Phase 1: RUL prediction using Transformer framework}
    Initialize the Transformer model with appropriate hyperparameters\;
    Process input sequence $\mathbf{X}$ through Transformer layers\;
    Generate RUL prediction $\boldsymbol{y} = (y_1, y_2, \ldots, y_m)$\;
    
    \tcc{Phase 2: Maintenance action recommendation using DRL}
    Initialize DRL model with neural network weights\;
    Use RUL prediction $\boldsymbol{y}$ as the observation state $\boldsymbol{s}$\;
    Select action $a$ based on $\epsilon$-greedy policy using state $\boldsymbol{s}$\;
    Execute action $a$ and observe new state and reward\;
    Update DRL model using observed reward and new state\;  
    Adjust $\epsilon$ for exploration-exploitation balance\;
    Output recommended action $a$\;
    \Return $a$
\end{algorithm}

\subsection{Case 1: Remaining Useful Life Prediction Module for a Single Machine}

Transformers, originating from natural language processing~\cite{wen2022transformers}, have become valuable tools in time-series data analysis, due to their proficiency in understanding complex temporal sequences. Our proposed Algorithm~\ref{algorithm-enhanced-transformer} leverages a transformer to forecast the RUL from historical time-series sensor data. The core of this approach is the self-attention mechanism, which establishes causal relations between different time steps based on query-and-key similarities, adapting to the complexity of the data, for example, self-attention can handle complex relationships in sensor data while issues depend on small changes over time in predictive maintenance.

\subsubsection{Encoder Module}
For a given time-series $ \mathbf{X} = [\boldsymbol{x}_1, \boldsymbol{x}_2, \cdots, \boldsymbol{x}_n]$, where $n$ is the sequence length and $\boldsymbol{x}_i \in \mathbb{R}^{d_{\text{model}}}$, the Encoder module converts $\mathbf{X}$ into hidden states $ \boldsymbol{h} = [h_1, h_2, \cdots, h_n ]$, with each $h_i \in \mathbb{R}^{d_{\text{model}}}$. This module comprises an input embedding layer and a number of  $L$ identical self-attention blocks as follows:
\begin{itemize}
    \item Input Embedding. The input data in $\mathbf{X}$ are processed with a learned embedding layer and converted to vectors of dimension $d_{\text{model}}$. The embedding results are then added with a learned positional encoding of the same dimension to reflect the temporal difference between the input samples.
    \item Multi-head Self-attention Mechanism. Given the matrices for query $\mathbf{Q} \in \mathbb{R}^{n \times d_k}$, key $\mathbf{K} \in \mathbb{R}^{n \times d_k}$, and value $\mathbf{V} \in \mathbb{R}^{n \times d_v}$, respectively, the attention score is computed as follows:
    \begin{align}
        \label{eq_self_atten}
        \text{Attention}(\mathbf{Q}, \mathbf{K}, \mathbf{V}) = \text{Softmax} \left( \frac{\mathbf{Q}\mathbf{K}^{T}}{\sqrt{d_k}} \right) \mathbf{V},
    \end{align}
    with $ d_k $ being the dimension of queries and keys, and $d_v$ the dimension of values. Note that $\mathbf{Q}$, $\mathbf{K}$ and$\mathbf{V}$ are derived from the same embedding/input with linear projection. In (\ref{eq_self_atten}), we call the sample-wise computation between the $(i,j)$-th elements as the attention weight:
    \begin{align}
         \alpha_{ij} = \text{Softmax} \left( \frac{\mathbf{Q}_i\mathbf{K}_j^T}{\sqrt{d_k}} \right),
    \end{align}
    since $\alpha_{ij} $ signifies the attention weight of the $j$-th input for the $i$-th output. With multiple heads, the queries, keys, and values are learned in parallel linear projections, and then concatenated as the output of this layer.
    
    \item Feed-Forward Neural Network (FFN). A single-layer FFN is utilized to capture non-linear patterns. This layer is represented as:
    \begin{align}
        \label{eq_FFN}
        FFN(\boldsymbol{x}_i) = \max(0, \boldsymbol{x}_i\mathbf{W}_1 + b_1)\mathbf{W}_2 + b_2,
    \end{align}
    with weight matrices $ \mathbf{W}_1, \mathbf{W}_2 $ and bias terms $ b_1, b_2 $.
\end{itemize}

\subsubsection{Decoder Module}

After encoding the time-series data into hidden states, the decoder further refines this information to produce outputs that are conducive to RUL prediction. Given hidden states $\boldsymbol{h}' $, the decoder module generates output states $\textbf{O} = \{\boldsymbol{o}_1, \boldsymbol{o}_2, \cdots, \boldsymbol{o}_n\}$. Similar to the encoder, the decoder module is also composed of a number of $L$ identical blocks. Each block in the decoder encompasses:

\begin{itemize}
\item Masked Multi-head Self-attention Mechanism. This layer takes the output embedding as the input and is the same as the multi-head self-attention mechanism in the encoder module. Masking is applied to make sure that the newly generated output only depends on the historical outputs.
\item Multi-head Cross-Attention Mechanism. Cross-attention takes in the encoder's outputs and combines them with the decoder's states. This process is represented as
\begin{align}
    \text{Attention}(\mathbf{Q}_d, \mathbf{K}_e, \mathbf{V}_e) = \text{Softmax} \left( \frac{\mathbf{Q}_d \mathbf{K}_e^\mathbf{T}}{\sqrt{d_k}} \right) \mathbf{V}_e,
\end{align}
where $ d_k $ represents the dimension of the queries and keys. Being different from (\ref{eq_self_atten}), the values of $\mathbf{K}_e$ and $\mathbf{V}_e$ are derived from the output of the encoder and $\mathbf{Q}_d$ is derived from the output embedding of the decoder or the output of the previous block in the decoder.
\item Feed-forward Neural Network. This module extracts patterns in the decoder states and takes the same form as (\ref{eq_FFN}) in the encoder:
\[ FFN(\boldsymbol{o}_i) = \max(0, \boldsymbol{o}_i\mathbf{W}_3 + b_3)\mathbf{W}_4 + b_4, \]
with weight matrices $ \mathbf{W}_3, \mathbf{W}_4 $ and biases $ b_3, b_4 $.
\end{itemize}

We note that each attention layer or FFN layer in the encoder and the decoder is accompanied with a residual connection and their output is processed with layer normalization (see Algorithm~\ref{algorithm-enhanced-transformer}).

\subsubsection{Prediction Module}
The prediction module utilizes the states $\boldsymbol{h}$ to estimate the RUL. This involves one or several dense FFN layers, each layer of which can be expressed as
\[ \text{RUL} = \sigma (\boldsymbol{h}\mathbf{W} + b), \]
where the weight matrix $\mathbf{W}$, bias $b$, and activation function $\sigma$ ensure that the RUL remains within the range.

\begin{algorithm}[!ht]
    \caption{A RUL prediction algorithm for a single machine}
    \label{algorithm-enhanced-transformer}
    \SetAlgoLined
    \KwIn{Input sequence $\mathbf{X} = (\boldsymbol{x}_1, \boldsymbol{x}_2, \ldots, \boldsymbol{x}_n)$}
    \KwOut{Output sequence $\mathbf{y} = (y_1, y_2, \ldots, y_m)$}
    Initialize the positional encoding\;
    Initialize the input embeddings\;
    \For{$i \leftarrow 1$ \KwTo $L$}{
        \tcc{Encoder Layer $i$}
        Apply multi-head self-attention to the input sequence\;
        Add residual connection followed by layer normalization\;
        Apply position-wise feed-forward networks\;
        Add another residual connection followed by layer normalization\;
    }
    \tcc{Decoder}
    \If{task requires a decoder}{
        \For{$j \leftarrow 1$ \KwTo $L$}{
            \tcc{Decoder Layer $j$}
            Apply masked multi-head self-attention to the output of the encoder\;
            Add residual connection followed by layer normalization\;
            Apply multi-head attention combining encoder outputs and decoder inputs\;
            Add residual connection followed by layer normalization\;
            Apply position-wise feed-forward networks\;
            Add another residual connection followed by layer normalization\;
        }
    }
    Generate output sequence $\mathbf{y}$ based on the final layer's output\;
    \Return $\mathbf{y}$
\end{algorithm}

\begin{algorithm}[!ht]
    \caption{A federated RUL prediction algorithm for multiple machines}
    \label{algorithm-fl-transformer}
    \SetAlgoLined
    \KwIn{Distributed input sequences $\mathbf{x}_1, \mathbf{x}_2, \ldots, \mathbf{x}_K$ from $K$ machines}
    \KwOut{Aggregated output sequences $\mathbf{y}_1, \mathbf{y}_2, \ldots, \mathbf{y}_K$}

    Initialize a central model with positional encoding and input embeddings;

    \For{each training round}{
        \tcc{Local Training on Each Machine}
        \For{each machine $k = 1$ \KwTo $K$}{
            Receive model parameters from the central model\;
            Apply Algorithm~\ref{algorithm-enhanced-transformer} with the given number of layers on local sequence $\mathbf{x}_k$\;
            Compute local updates and send them back to the central model\;
        }
        
        \tcc{Aggregation at the Central Model}
        Aggregate local updates from all machines\;
        Update central model parameters\;
    }

    \tcc{Output Generation}
    \For{each machine $k = 1$ \KwTo $K$}{
        Send final model parameters to machine $k$\;
        Generate output sequence $\mathbf{y}_k$ using the final model on local input $\mathbf{x}_k$\;
    }
    
    \Return $\mathbf{y}_1, \mathbf{y}_2, \ldots, \mathbf{y}_K$.
\end{algorithm}

\subsection{Case 2: A Federated RUL Prediction Module for Multi-Machines}
\label{subsec_fed_RUL}

The multi-machine case is featured by a federated model, which dispatches an individual training model for $K$ similar machines. A local model $k$ takes as input the local time series data $\mathbf{X}_k$ and outputs an estimation of the RUL $R_k(t)$ regarding the machine it concerns. With sensors deployed in the same way, different local models observe time-series inputs of the same dimension and share the same prediction label set. By combining the strengths of multiple models extracted from different machines, the approach improves the reliability and accuracy of RUL predictions, enabling industries to make informed decisions regarding maintenance schedules and machinery operations. The global model for RUL prediction leverages an ensemble of predictive models to provide a more precise estimate of RUL, enhancing the efficiency and reliability of predictive maintenance in industries.
 
As shown in the Algorithm~\ref{algorithm-fl-transformer}, given a set of time series data $\mathbf{X}$, the ensemble model aims to predict the RUL $R(t)$ of multiple machines. Formally, the ensemble model can be defined as a function:
\begin{align}
    R(t) = f_{\text{ensemble}}(\mathbf{X}),
\end{align}
where $f_{\text{ensemble}}$ is the ensemble model that aggregates the predictions of multiple individual models.

By integrating the predictions of multiple models, the ensemble approach refines the reliability and accuracy of the RUL predictions. This enhanced prediction quality allows industries to make precise maintenance schedules and optimize machine operations. The global model for RUL prediction can be represented as
\begin{align}
    R(t) = \frac{1}{K} \sum_{k=1}^{K} f_k(\mathbf{X}),
\end{align}
where $K$ is the number of models in the ensemble, and $f_k$ represents the individual predictive model. This averaging method capitalizes on the collective strength of diverse predictive algorithms. Based on the predictive RULs, the infrastructure can conduct equipment maintenance or replacement to ensure uninterrupted and reliable functionality of the machines.

The distinction between Case 1 and Case 2 lies primarily in their scale and data handling methods. Case 1 focuses on predicting the RUL for a single machine using a Transformer model, which excels in interpreting complex temporal sequences in sensor data. This approach is highly specialized for analyzing the intricate patterns of an individual machine. In contrast, Case 2 extends to multiple machines, employing a federated learning approach with an ensemble model. This method aggregates data from various machines, synthesizing diverse inputs for a broader analysis. It suits an industry that contains multiple similar production lines or machines. Additionally, Case 2 can suitably apply Case 1's algorithm to each machine and then focus on multi-machine data aggregation.

\begin{algorithm}[!ht]
    \caption{A DRL-based algorithm for optimal maintenance action recommendation}
    \label{algorithm-generalized-RL}
    \SetAlgoLined
    \KwIn{Exploration rate $\epsilon$ (for DQN, SAC), learning rate $\alpha$, discount factor $\gamma$ and policy update frequency (for PPO, SAC).}
    \KwOut{Optimal action $a$ for maintenance}

    \tcc{Initialization}
    Initialize the policy network and optionally value/critic network(s) with the input hyperparameters\;
    Initialize target network(s) if applicable (e.g., for DQN, SAC)\;
    Initialize experience replay buffer if applicable\;

    \For{each episode}{
        Initialize the starting state $\boldsymbol{s}$\;
        \While{$\boldsymbol{s}$ is not a terminal state}{
            \tcc{Action Selection}
            \uIf{using DQN or SAC}{
                Select action $a$ using an $\epsilon$-greedy policy or according to the current policy and exploration strategy\;
            }
            \ElseIf{using PPO}{
                Select action $a$ according to the current policy\;
            }
            Execute action $a$, observe reward $r$ and successor state $\boldsymbol{s'}$\;

            \tcc{Policy and (or) Value Network Update}
            \uIf{using DQN}{
                Store transition $(\boldsymbol{s}, a, r, \boldsymbol{s'})$ in replay buffer\;
                Sample mini-batch and update the network(s) using TD error\;
                Periodically update target network\;
            }
            \uElseIf{using PPO}{
                Accumulate data for policy update\;
                Update policy using PPO\;
            }
            \ElseIf{using SAC}{
                Store transition in replay buffer\;
                Sample mini-batch and update policy and value networks using SAC update rules\;
                Adjust temperature parameter for entropy maximization\;
            }

            \tcc{Exploration Rate Adjustment}
            Adjust the exploration rate $\epsilon$ if applicable\;
            Update current state $\boldsymbol{s} \leftarrow \boldsymbol{s'}$\;
        }
        \tcc{Policy Evaluation}
        Periodically evaluate the performance of the current policy\;
    }
    \Return Optimal maintenance action $a$
\end{algorithm}

\subsection{Deep Reinforcement Learning for Maintenance Action Recommendation}

Upon obtaining the RUL predictions from the previous module, the maintenance action recommendation module initiates. This module uses the predicted RUL as a primary determinant, facilitating the calibration of the urgency, type, and extent of necessary maintenance interventions. If the predicted RUL of a machine or component is longer, there is less need for urgent maintenance. In contrast, if the predicted RUL is shorter, it is important to quickly perform maintenance actions or repairs to prevent problems. Thus, we select the most important part of RUL using principal component analysis (PCA) as follows:

\textbf{Rule-Based Systems.} These systems are established on the basis of predetermined heuristic principles. For instance, should the RUL prediction for a component fall below a designed limit (e.g., 10 hours), an immediate maintenance notification is dispatched. On the other hand, if the RUL surpasses the limit, periodic inspections are considered sufficient.

\textbf{Strategy-Learning Agents.} These agents are trained using historical datasets, where earlier RUL predictions are coupled with their corresponding effective maintenance interventions. Inputting the present RUL prediction enables the agent to recommend the most probable efficient maintenance actions, grounded in historical insights.

The appropriate integration of the RUL predictions with the maintenance action recommendation module ensures that maintenance schedules are efficient and data-driven, ultimately leading to improved system reliability, longer equipment lifespan, and reduced operational costs. Thus, the predicted RUL, indicative of a system's operational state, is assimilated into the state representation within a Markov Decision Process (MDP)-based framework. The subsequent discussions elaborate on the maintenance strategy-learning procedure under the MDP framework, which includes the following components:

\textbf{Environment State Space.} As established, the system can be in one of $ n $ distinct degradation states, with $ \mathbf{S}^t = \{S_1^t, S_2^t, \ldots, S_n^t\} $ consisting all possible $n$ states that the system can take at time $ t $.

 \textbf{Action Space.} If we denote the number of possible maintenance actions by $ m $, the action space $ \mathcal{A} $ can be represented as $ \mathcal{A} = \{a_1, a_2, \ldots, a_m\} $.

\textbf{State Transition Probabilities.} The probability of transitioning from state $ s_i^t $ to state $ s_j^{t+1} $ given action $ a_k^t $ is represented by
$ P_{i,j}^k = P(s_{t+1} = s_j^{t+1} | s_t = s_i^t, a_t = a_k^t),$ where the complete state transition probability matrix given action $ a_k^t $ would then be a matrix $ \mathbf{P}^k $ where each entry $ P_{i,j}^k $ represents the transition probability from state $ s_i^t $ to state $ s_j^{t+1} $.

 \textbf{Reward Structure.} The immediate reward received after transitioning from state $ s_i^t $ to $ s_j^{t+1} $ upon taking action $ a_k^t $ is given by $ R(s_i^t, a_k^t, s_j^{t+1}) $. The expected reward for taking action $ a_k^t $ in state $ s_i^t $ is:
\[ R(s_i^t, a_k^t) = \sum_{j=1}^n P_{i,j}^k R(s_i^t, a_k^t, s_j^{t+1}). \]

\textbf{Policy.} A policy $ \pi $ is a (probabilistic) mapping from states to actions. An optimal policy $ \pi^* $ maximizes the expected cumulative discounted reward from any starting state. The value of a state $ s_i $ under policy $ \pi $ is:
\[ V^\pi(s_i) = R(s_i, \pi(s_i)) + \gamma \sum_{j=1}^n P_{i,j}^{\pi(s_i)} V^\pi(s_j), \]
where $ \gamma $ is the discount factor for future rewards.

\textbf{Optimal Policies under Bellman Equation.} The value of a state under an optimal policy $ \pi^* $ obeys the following recursive Bellman optimality equation
\begin{align}
    V^*(s_i) = \max_{a \in A} \left[ R(s_i, a) + \gamma \sum_{j=1}^n P_{i,j}^a V^*(s_j)\right].
\end{align} 
On top of this MDP-based policy derivation model, we can develop a DRL-based plug-and-play module for maintenance action recommendation using neural network-based function/policy approximation, by accommodating many state-of-the-art DRL algorithms.
Algorithm~\ref{algorithm-generalized-RL} shows details of leveraging different DRL algorithms, including DQN, SAC, and PPO (see our later discussion for the details of each algorithm), for selecting optimal maintenance action $a$ through evaluating the performance of each policy.

In the RUL prediction stage, experiment results in Section~\ref{sec:case} show that our proposed framework succeeds in predicting RULs using four sub-datasets in the NASA C-MAPSS open dataset~\cite{saxena2008damage}. However, a gap still exists between the prediction results and the maintenance action decisions needed by engineers. To make action recommendations by weighing the cost and benefits, we leverage DRL algorithms as given in Algorithm~\ref{algorithm-generalized-RL}, which helps to find the best trade-off.

\subsubsection{States and Actions}

The state space is generated from PCA features of the RUL resulting from the prediction outcomes of NASA's C-MAPSS Aircraft Engine Simulator dataset. These PCA features provide a compact and important part of the current health condition of the machine. The action space contains three actions: 
\begin{itemize}
    \item No Action ($a_1$), appropriate for healthy machine states, minimizes immediate costs but risks penalties in critical conditions;
    \item Partial Maintenance ($a_2$), suitable for moderate unhealthy states, involves preventive actions like lubrication to extend machine life at moderate costs;
    \item Complete Overhaul ($a_3$), necessary for critical machine states, entails comprehensive repairs and replacements to prevent catastrophic failures but incurs high costs and downtime.
\end{itemize}
The DRL agent aims to navigate this state-action space to minimize long-term operational costs by selecting the optimal maintenance actions based on each state's representation of machine health.

\subsubsection{Probabilities Model}

For the purpose of evaluating the performance of our proposed algorithm, the transition and action cost matrices are empirically calibrated using the NASA C-MAPSS dataset to generate the controllable ground truth. Specifically, each entry in the state transition matrix $p(s_{t+1} | s_t, a_t)$ reflects the probability of transitioning from state $s_t$ to $s_{t+1}$ when action $a_t$ is taken, and these probabilities are derived through statistical analyses of state sequences and their corresponding RUL values in the C-MAPSS dataset. Moreover, action-related costs, such as maintenance expenses or energy consumption, are grounded in real-world data from this dataset. In this MDP framework, the transition matrix has a dimension $(10, 3, 10)$, representing transitions between $10$ discretized states given $3$ actions, all closely reflecting scenarios that commercial aero-engines might encounter.

\subsubsection{Reward Model}

The objective of maintenance actions recommends algorithms not only to maximize the RUL but also to minimize operational costs and downtime. To fulfill these multi-objective requirements, we propose a reward function that combines these elements in a weighted manner. The reward function $ R(s_t, a_t, s_{t+1})$ is formulated as
\begin{align}
 \label{eq_original_reward}
 R(s_t, a_t, s_{t+1}) &= \alpha \times \text{RUL}_{\text{gain}}(s_t, a_t)~\nonumber\\
 &- \beta \times \text{Cost}(a_t) - \gamma \times \text{DownTime}(a_t),
\end{align}
where
\begin{itemize}
    \item $\text{RUL}_{\text{gain}}(s_t, a_t)$ represents the gain in RUL as a consequence of taking action $a_t$ at state $s_t$. This gain is computed using the state-transition probabilities $p(s_{t+1}| s_t, a_t)$ derived from the C-MAPSS dataset. A higher RUL gain is desirable as it suggests an extension in the machine's normal working time.
    \item $\text{Cost}(a_t)$ denotes the operational or maintenance cost incurred by taking action $a_t$. This cost is modeled based on experience.
    \item $\text{DownTime}(a_t)$ signifies the machine downtime resulting from action $ a_t $. Downtime is critical in operational settings, which impacts productivity or service levels.
    \item $\alpha, \beta, \text{and}~\gamma$ are hyperparameters that weigh the importance of RUL, cost, and downtime, respectively. By fine-tuning these hyperparameters, the model can be customized to focus on extending the RUL, minimizing costs, or reducing downtime.
\end{itemize}

\subsection{Integrate with Human Feedback for Maintenance Action Recommendation}

Incorporating human expertise into the reinforcement learning process helps to resolve some of the challenges encountered with traditional RL methods. Especially, in scenarios where real-world rewards are hard to quantify or can be misleading, human feedback provides a way to guide the learning process more effectively. Thus, as machines and their associated systems become increasingly complex, conditions that need to be considered may take a long time for Algorithm~\ref{algorithm-generalized-RL}.  Human experts, with their working experiences and domain knowledge to find complex patterns, can provide valuable insights that may not be readily available through raw data. To minimize labor costs, we assume that one expert will be involved during the human feedback process. By integrating human feedback into the RL process, the framework can:
\begin{itemize}
    \item Augment the learning process with domain-specific knowledge.
    \item Address the cold start problem where initial exploration might be risky or expensive.
    \item Improve learning efficiency by reducing the exploration of sub-optimal actions.
\end{itemize}

The above goals are achieved by incorporating the following information into our proposed framework in Algorithm~\ref{algorithm-rlhf-revised}:
\begin{enumerate}
    \item \textit{Human Opinions}: Collect human predictions of the desired behavior. Experienced engineers can provide opinions on maintenance decisions based on various machine states obtained from the C-MAPSS dataset.
    \item \textit{Feedback on Agent's Actions}: After the agent makes a decision, a human can provide feedback on the chosen action, guiding the agent on the appropriate actions.
\end{enumerate}

Then, we introduce a term in the reward function that captures human feedback
\begin{align}
   R_{\text{human}}(a_t) = 
\begin{cases} 
r_{\text{positive}}, & \text{if human feedback is positive}, \\
r_{\text{negative}}, & \text{if human feedback is negative}, \\
0, & \text{otherwise},
\end{cases} 
\end{align}
based on which, the modified reward function from (\ref{eq_original_reward}) for DRL becomes
\begin{align}
 \nonumber   R(S_t, a_t, S_{t+1}) &= \alpha \times \text{RUL}_{\text{gain}}(S_t, a_t) - \beta \times \text{Cost}(a_t) - \\&\gamma \times \text{DownTime}(a_t) \nonumber + \delta \times R_{\text{human}}(a_t), \nonumber
\end{align}
where \(\delta\) is a hyperparameter weighing the importance of human feedback.

\begin{algorithm}[!t]
    \caption{A RLHF-based algorithm for optimal maintenance action recommendation}
    \label{algorithm-rlhf-revised}
    \SetAlgoLined
    \KwIn{Observation $\boldsymbol{s}$, Exploration rate $\epsilon$, Learning rate $\alpha$, Discount factor $\gamma$, Human feedback $0$.}
    \KwOut{Optimal action $a$ for maintenance}

    Initialize the neural network weights\;
    Initialize a mechanism to incorporate human feedback (e.g., reward shaping, action recommendation)\;
    
    \For{each episode}{
        \tcc{Incorporate Human Demonstrations}
        If available, utilize human demonstrations to set initial policy\;

        Initialize the starting state $\boldsymbol{s}$\;
        \While{$s$ is not a terminal state}{
            \tcc{Action Selection with Human Feedback}
            Select action $a$ using an $\epsilon$-greedy policy or based on human recommendations\;
            Execute action $a$, observe reward $r$ and successor state $\boldsymbol{s'}$\;
            
            \tcc{Adjust Reward Based on Human Feedback}
            Incorporate human feedback into the reward:
            \[
            R_{\text{human}}(a_t) = 
            \begin{cases} 
            r_{\text{positive}}, & \text{if HF is positive}, \\
            r_{\text{negative}}, & \text{if HF is negative}, \\
            0, & \text{otherwise}.
            \end{cases} 
            \]
            Adjust reward $r$ based on $R_{\text{human}}(a_t)$\;

            \tcc{Network Update}
            Calculate TD error $\delta = r + \gamma \max_{a'}Q(\boldsymbol{s'}, a') - Q(\boldsymbol{s}, a)$\;
            Update neural network weights using $\delta$ and learning rate $\alpha$\;

            \tcc{Adjustment of Exploration and State}
            Adjust the exploration rate $\epsilon$\;
            Update current state $\boldsymbol{s} \leftarrow \boldsymbol{s'}$\;
        }
        \tcc{Incorporate Feedback into Learning Process}
        Periodically adjust the learning process based on accumulated human feedback\;
    }
    \Return Optimal maintenance action $a$.
\end{algorithm}

\section{Case Study}
\label{sec:case}

In this section, we present a case study of our proposed TranDRL framework.
\subsection{Data}
NASA's C-MAPSS dataset~\cite{saxena2008damage}  contains four subsets denoted as FD001, FD002, FD003, and FD004, each with a training and test set, reflecting distinct operating and fault modes. The training data comprise inputs from $3$ operating conditions and $21$ sensor measurements across four engines of the same type. While the datasets are i.i.d, each engine has its unique operational settings and sensor readings. By leveraging the C-MAPSS dataset, we can gain insights into how IoT systems might perform in real-world scenarios, particularly those involving complex machinery such as aircraft engines. The dataset facilitates data collection and monitoring, mimicking real-world engine performance, which represents the general working scenario for IoT-based sensing and decision making in different industrial sectors. Therefore, it provides a convenient platform for the development and testing of predictive maintenance and anomaly detection algorithms by offering simulated engine degradation and fault data.

\subsection{Experiment Setup}

In the experiment, we applied a piecewise linear degradation model~\cite{Heimes2008RecurrentNN} for RUL determinations. Each engine's last cycle data point generally stands as the sample, with actual RUL values included in the dataset.

\textbf{Environment Settings.} We utilized an Intel Core i9-14900K CPU processor, 128GB DDR5 RAM, and a PNY RTX 4090 GPU card. The system ran on Ubuntu 22.04 LTS with Python 3.9 and PyTorch 2.0.1 for tasks.

\textbf{Hyperparamters.} We carefully choose its corresponding hyperparameters based on empirical experiments. 
    
\textit{Transformers.} The TFT model employs a learning rate of 0.03, hidden layers of size 16, a single attention head, a dropout rate of 0.1, a hidden continuous size of 8, and Symmetric Mean Absolute Percentage Error (SMAPE) as the loss function, focusing on the target variable RUL. We also introduce two baseline algorithms for prediction performance comparison. For the first reference algorithm, the N-Beats model~\cite{oreshkin2019n}, also set with a learning rate of 0.03, uses an encoder length of 120, a prediction length of 60, a batch size of 128, implement early stopping at 10 epochs, and uses gradient clipping at 0.1. For another reference algorithm, the DeepAR model~\cite{salinas2020deepar}, parameters include a lower learning rate of 0.003, encoder and decoder lengths of 120 and 60 respectively, the same batch size of 128, target normalization via GroupNormalizer, training up to 10 epochs with early stopping after 4, and weight decay at 0.01. The negative binomial distribution loss is employed.

\textit{Deep Q-network (DQN).}
DQN reduces the computational burden of tabular Q-learning by employing a deep neural network to approximate the state-action value function~\cite{fan2020theoretical}. It also introduces experience replay to improve learning stability and efficiency. The learning rate is set to $10^{-4}$, the discount factor is set to $0.99$, and the replay buffer size is set to $10^6$ experiences. We use a batch size of $256$ and update the target network every $50$ step. The exploration-exploitation trade-off is controlled by an $\epsilon$ value that started at $1.0$ and decayed over time to a final value of $0.01$. Additionally, we use a neural network with two hidden layers of $256$ units each, and we utilize the Adam optimizer with a learning rate of $10^{-4}$.

\textit{Proximal Policy Optimization (PPO).} 
PPO ensures more stable and efficient policy updates by employing a clipped objective function, thereby avoiding large, destabilizing updates to the policy~\cite{schulman2017proximal}. We set the learning rate for the actor and critic networks to $0.0001$ and the discount factor to $0.99$. A clip range of $0.2$ is used to limit the updates to the policy, and we apply a value function coefficient of $0.5$ to balance the importance of the policy and the value function. Additionally, we set the entropy coefficient to $0.01$ to encourage exploration and a batch size of $256$. Then, we optimize the policy every 10 epochs and use the Adam optimizer with a learning rate of $10^{-4}$. Finally, the GAE parameter is set to $0.95$ to estimate the advantage function.

\textit{Soft Actor-Critic (SAC).} 
SAC is particularly effective for problems with high-dimensional action spaces~\cite{haarnoja2018soft}. SAC aims to maximize an objective function that balances both the expected return and the entropy of the policy, making it more robust and efficient. The learning rate for the actor and critic networks is set to $10^{-4}$, and we use a discount factor of $0.99$. Then, we apply a temperature parameter $\alpha$ of $0.2$ to balance the entropy term in the objective function. The target entropy is set to $-2.0$, and we use a replay buffer of $10^6$ experiences with a batch size of $256$. We also apply a soft update of the target networks with a rate of $0.005$, and we initialize the log standard deviation of the policy with a value of $-2$. Finally, we train the algorithm for $10^5$ time steps and use the Adam optimizer with a learning rate of $10^{-4}$.

\begin{figure}[!ht]
     \centering
     \begin{subfigure}[b]{0.45\textwidth}
         \centering
         \includegraphics[width=\textwidth]{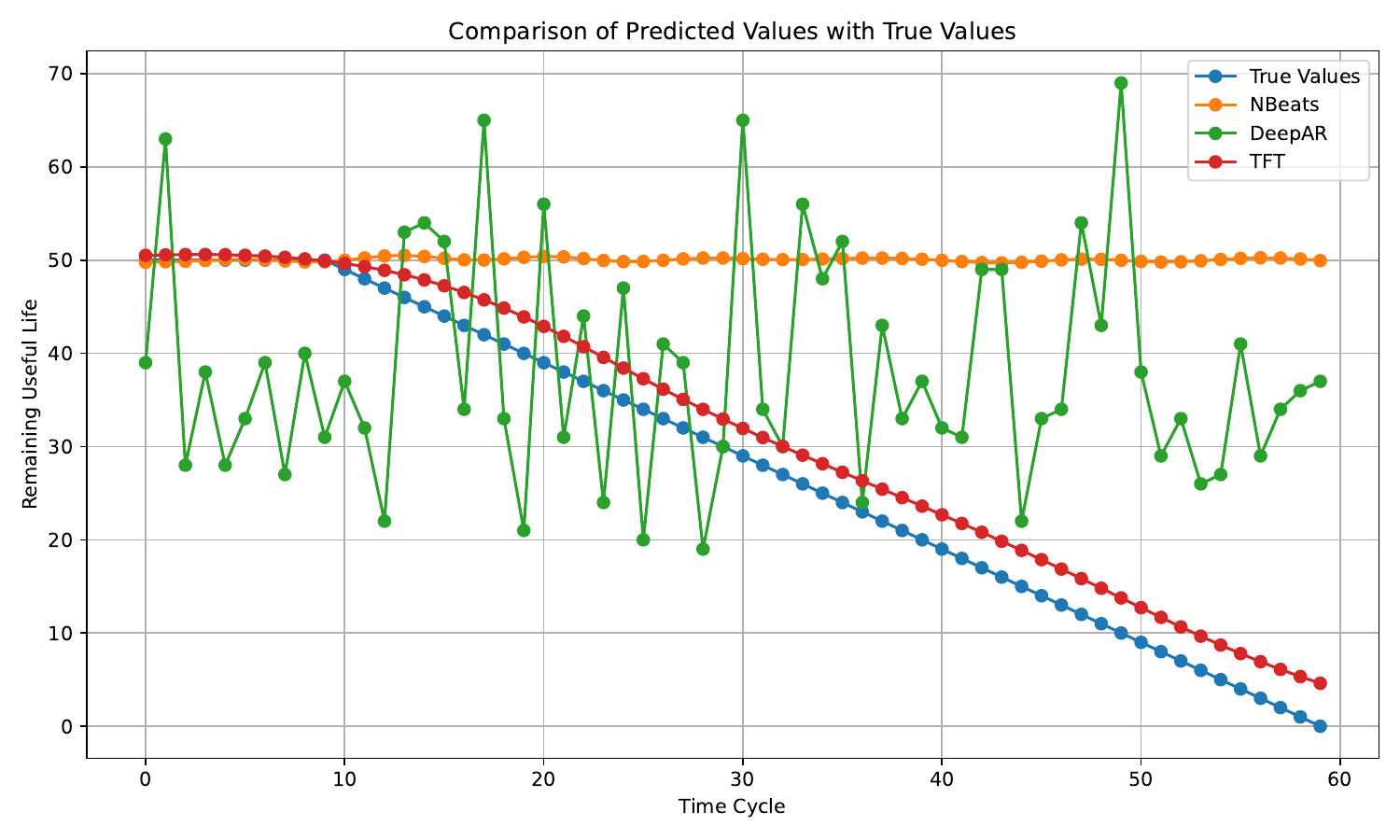}
         \caption{The RUL prediction results for machine FD001. }
         \label{fig-001}
     \end{subfigure}
     \hfill
     \begin{subfigure}[b]{0.45\textwidth}
         \centering
        \includegraphics[width=\textwidth]{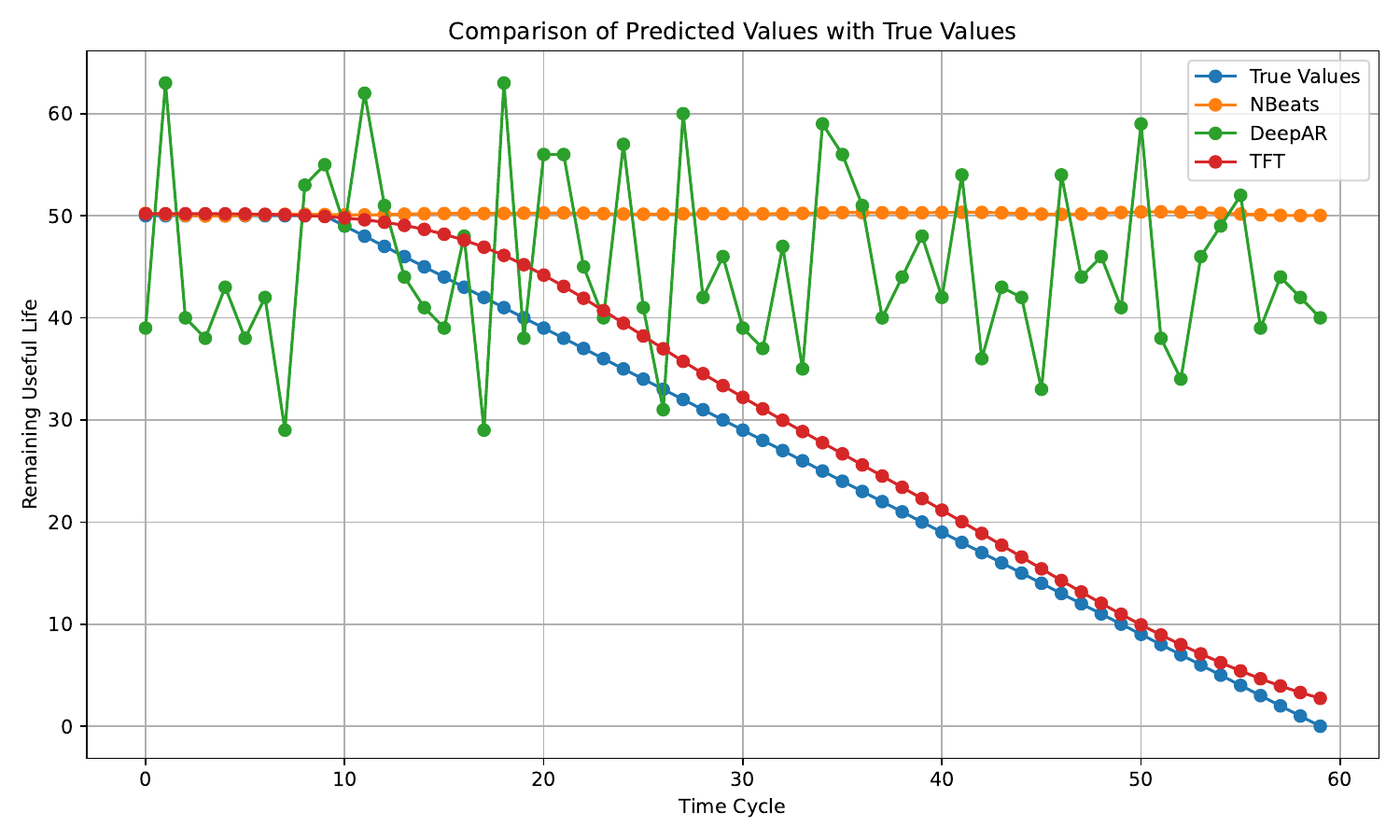}
        \caption{The RUL prediction results for machine FD002. }
        \label{fig-002}
     \end{subfigure}
     \hfill
     \begin{subfigure}[b]{0.45\textwidth}
         \centering
            \includegraphics[width=\textwidth]{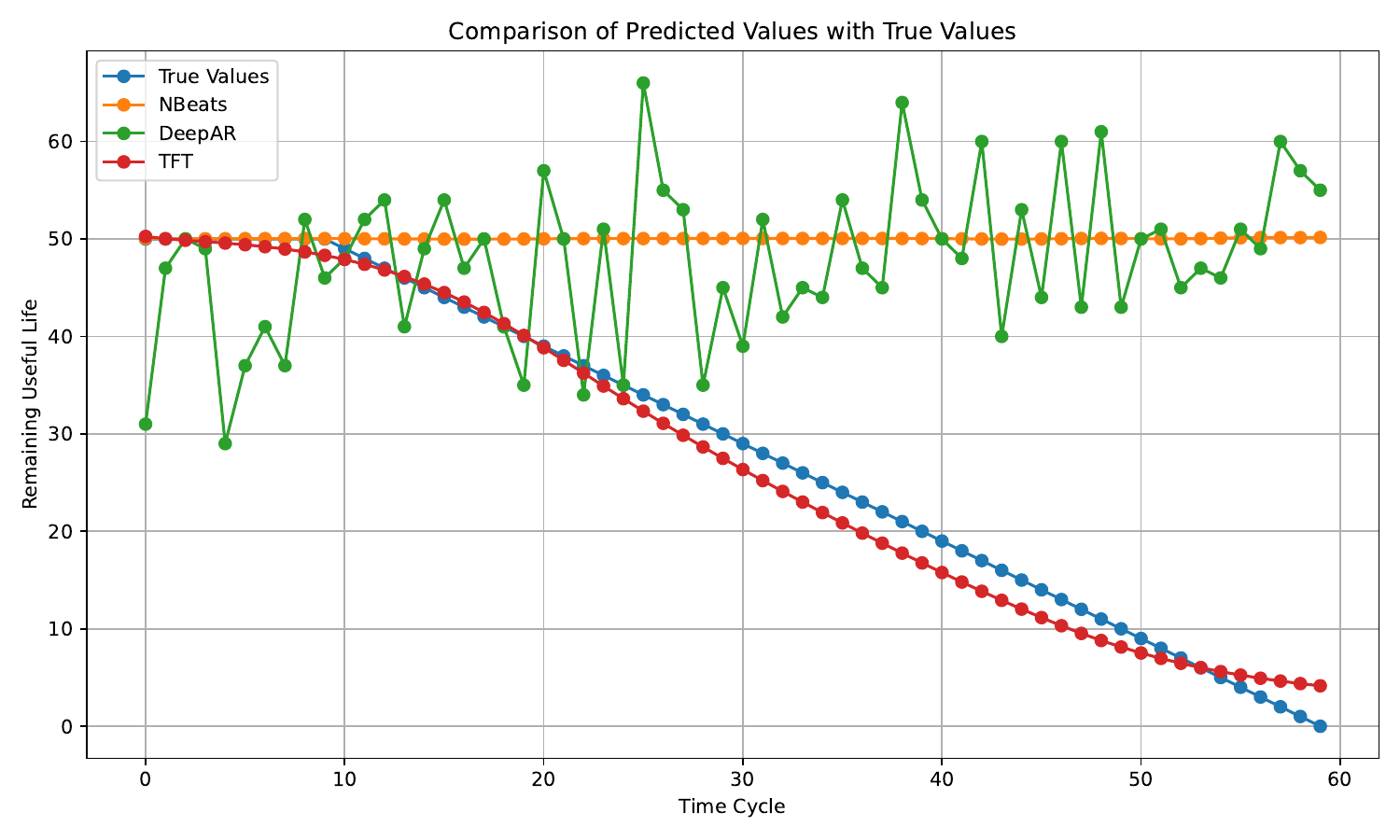}
            \caption{The RUL prediction results for machine FD003. }
            \label{fig-003}
     \end{subfigure}
     \hfill
     \begin{subfigure}[b]{0.45\textwidth}
        \includegraphics[width=\textwidth]{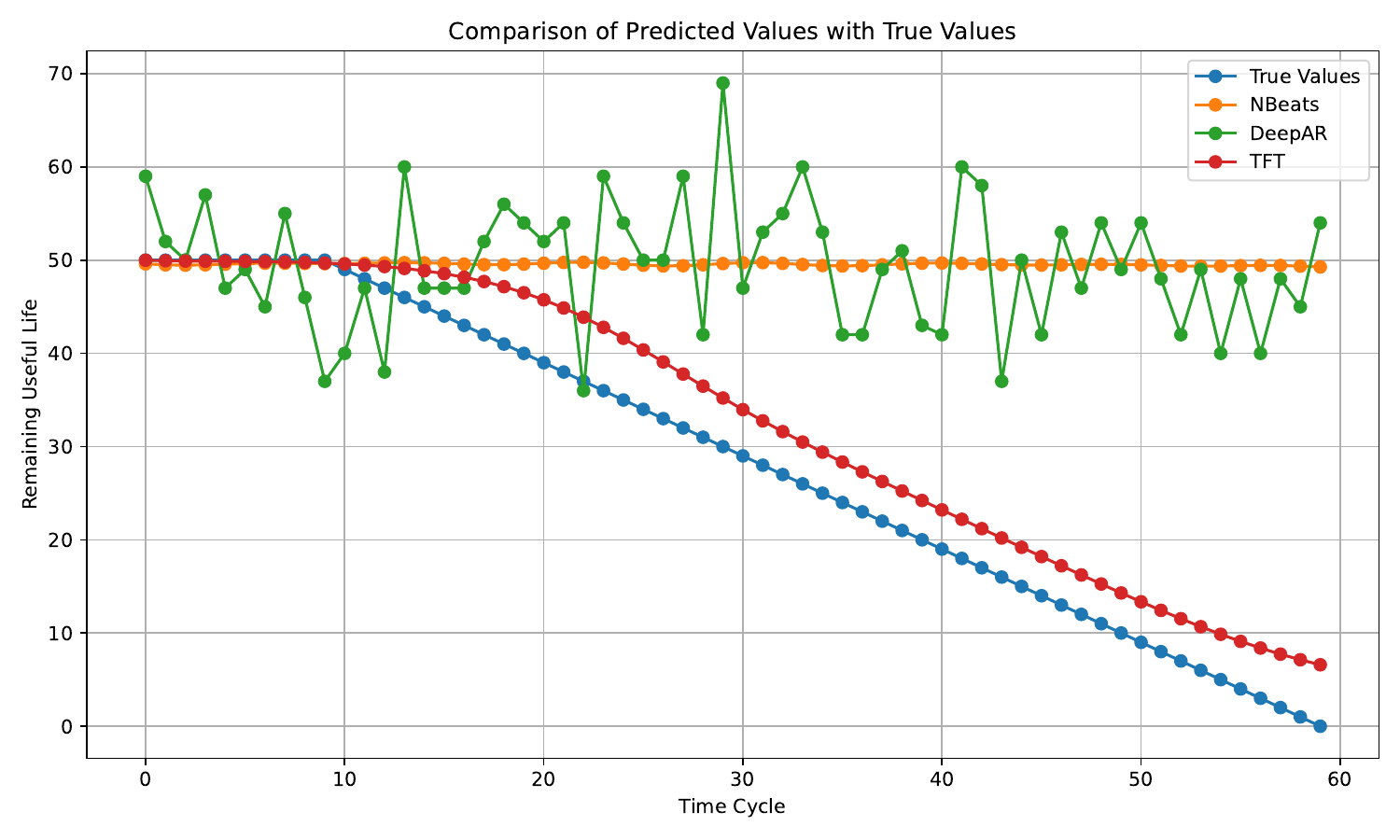}
        \caption{The RUL prediction results for machine FD004. }
        \label{fig-004}
         \end{subfigure}
    \caption{RUL prediction results for four machines.}
    \label{fig:main}
\end{figure}

\textbf{Baseline.} The ground truth is obtained from run-to-failure data, where the machine is operated until failure occurs. The run-to-failure data consists of time-series measurements of various sensor data recorded during operation until failure. The ground truth RUL values are calculated by subtracting the current cycle $t$ from the cycle $T_f$ at which failure occurred, expressed mathematically as $\text{RUL}(t) = T_f - t$. For baseline methods, we utilize NBeats~\cite{oreshkin2019n} and DeepAR~\cite{salinas2020deepar} to compare against our proposed method, aiming to verify whether it outperforms the baselines.

\subsection{Results and Analysis}

Figures~\ref{fig-001}-\ref{fig-004} demonstrate the superior performance of our proposed algorithm (Algorithm~\ref{algorithm-enhanced-transformer} as TFT) over NBeats and DeepAR in predicting the RUL for an individual machine, as evidenced by consistently higher  RUL values. This suggests TFT's enhanced capability in capturing complex, time-dependent patterns in high dimensional sensor data, highlighting its robustness across various machine behaviors.

\begin{figure*}
     \centering
     \begin{subfigure}[b]{0.45\textwidth}
         \centering
        \includegraphics[width=\linewidth]{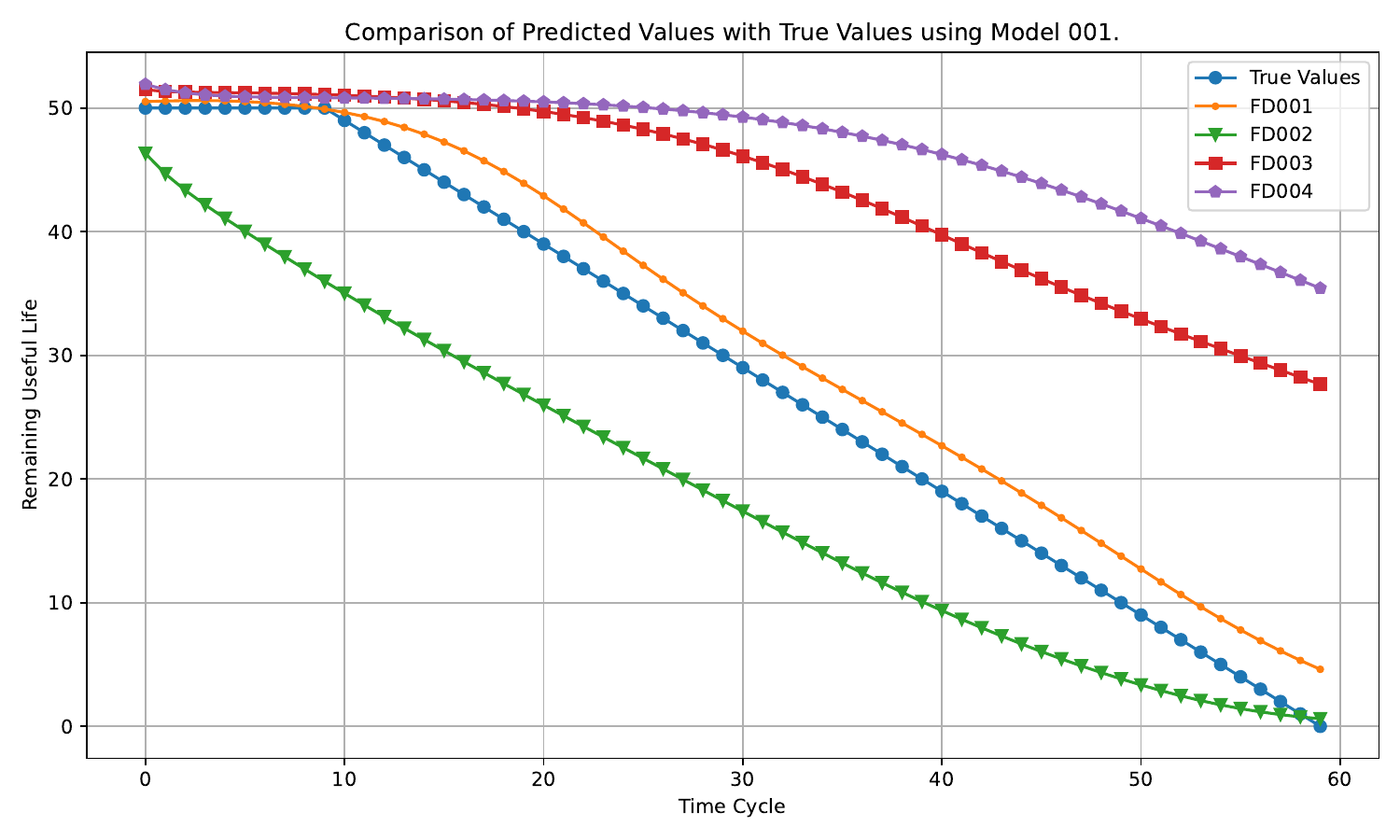}
        \caption{Comparison of RUL prediction results of four machines using the model trained from machine FD001. }
        \label{fig-model-001}
     \end{subfigure}
     \hfill
     \begin{subfigure}[b]{0.45\textwidth}
         \centering
        \includegraphics[width=\linewidth]{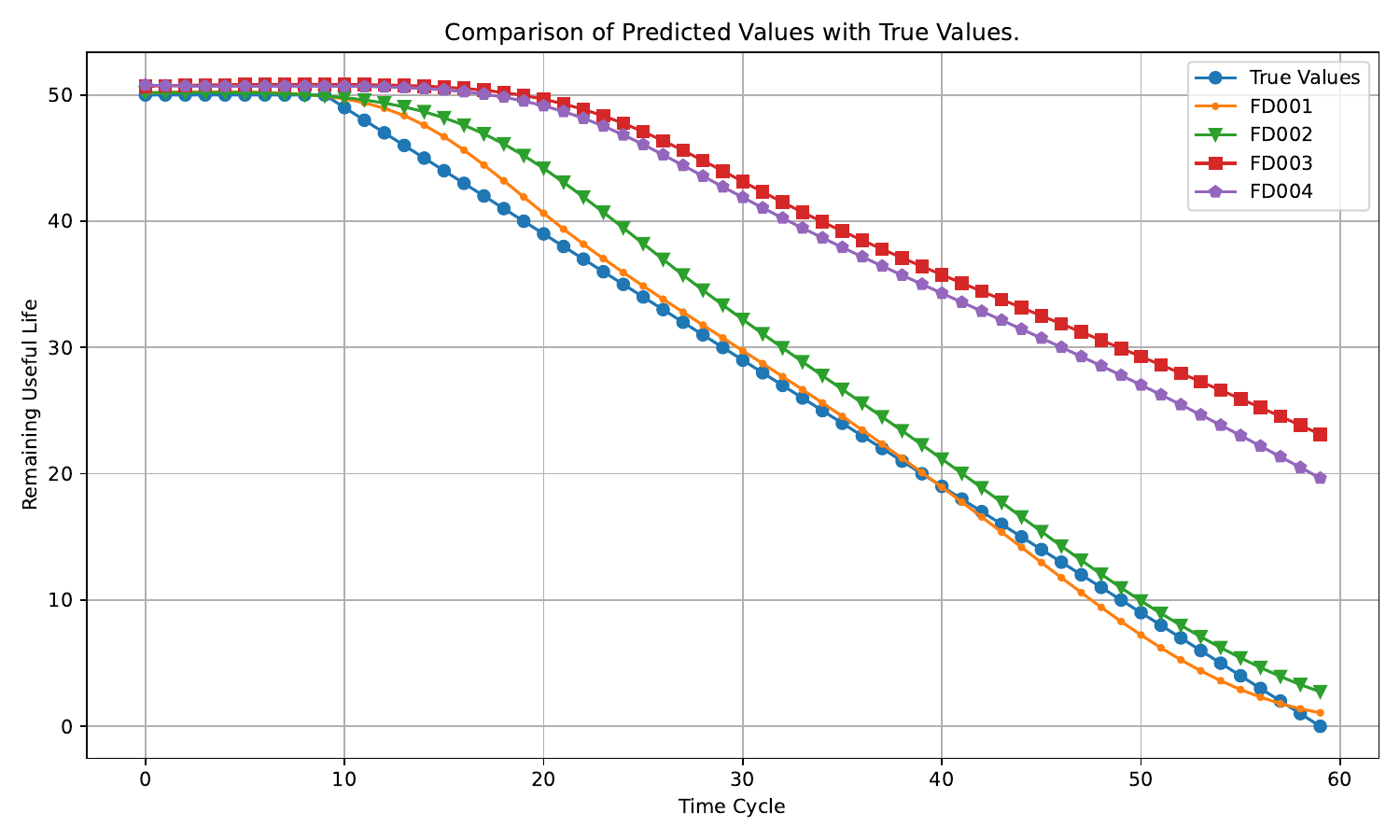}
        \caption{Comparison of RUL prediction results of four machines using the model trained from machine FD002. }
        \label{fig-model-002}
     \end{subfigure}
     \hfill
     \begin{subfigure}[b]{0.45\textwidth}
         \centering
        \includegraphics[width=\linewidth]{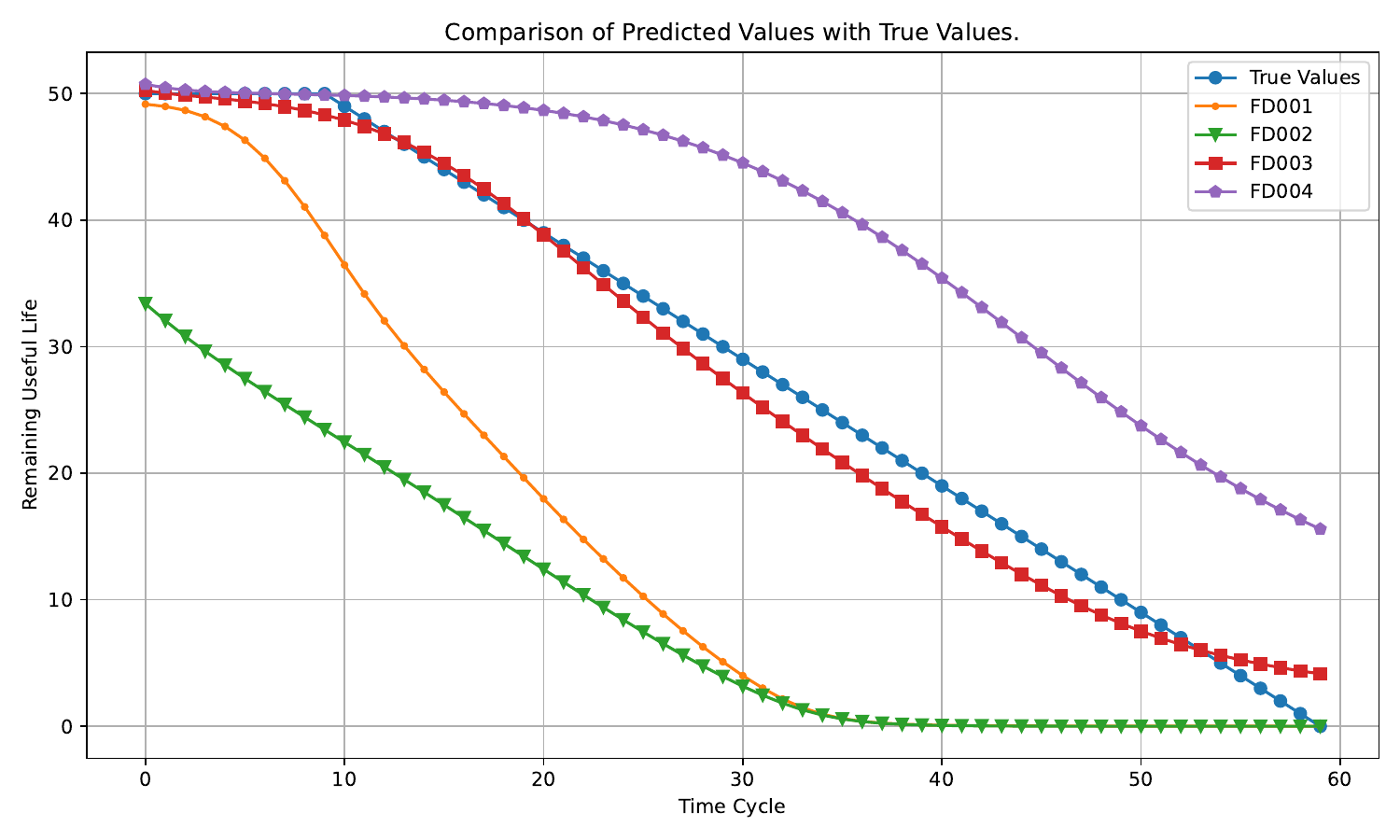}
        \caption{Comparison of RUL prediction results of four machines using the model trained from machine FD003.} 
        \label{fig-model-003}
     \end{subfigure}
     \hfill
     \begin{subfigure}[b]{0.45\textwidth}
        \includegraphics[width=\linewidth]{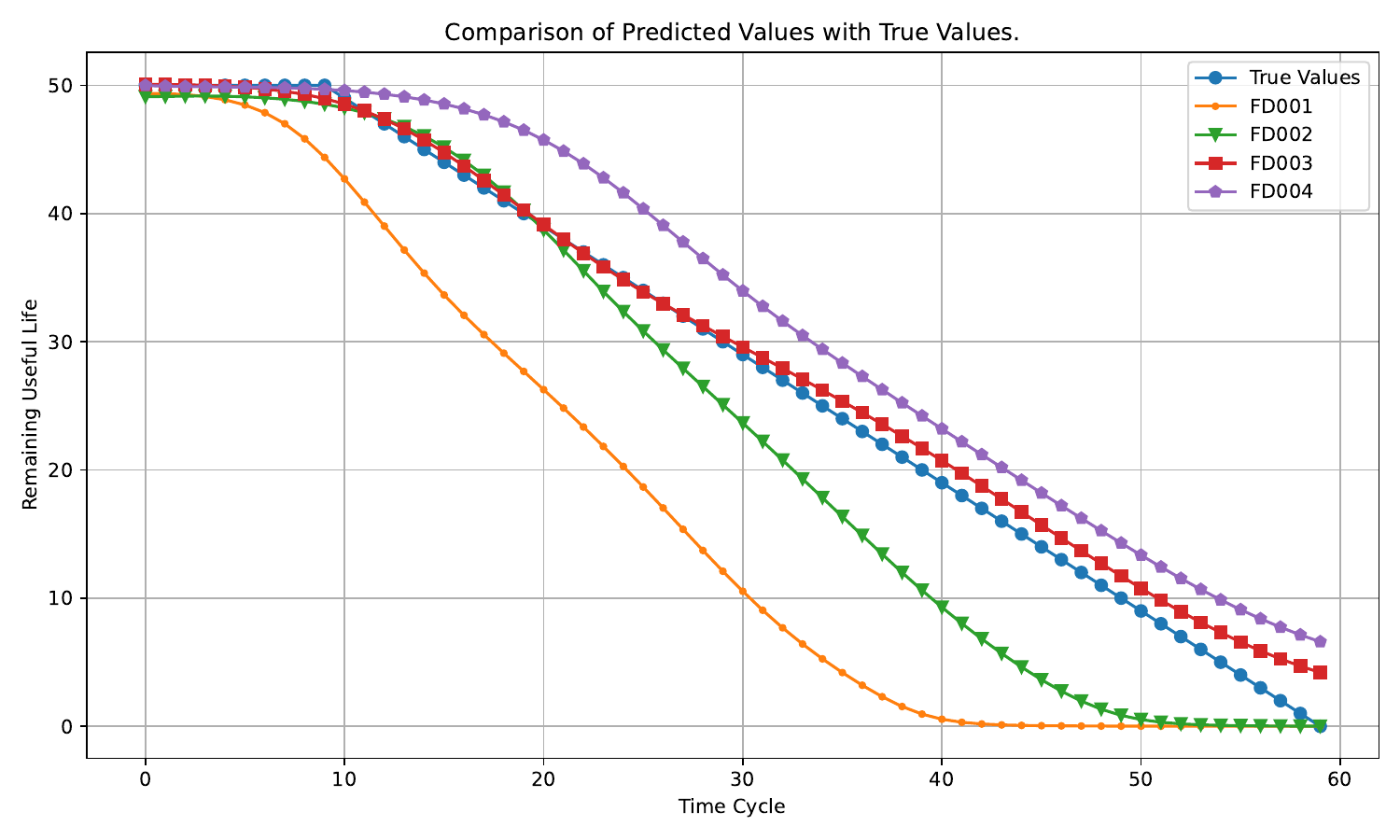}
        \caption{Comparison of RUL prediction results of four machines using the model trained from machine FD004.}
     \label{fig-model-004}
         \end{subfigure}
    \hfill
    \begin{subfigure}[b]{0.45\textwidth}
         \centering
         \includegraphics[width=\linewidth]{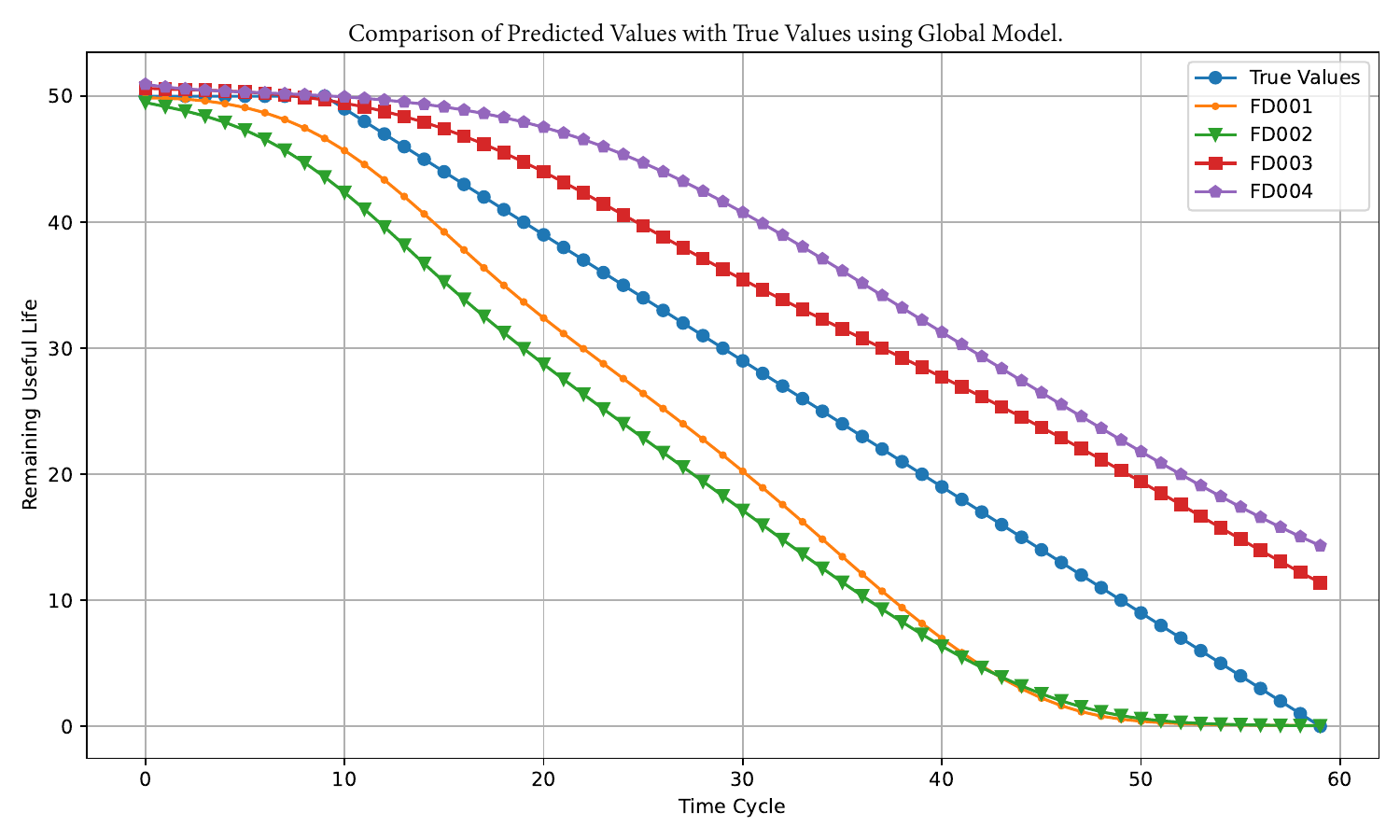}
        \caption{Comparison of RUL prediction results of four machines using the global model.}
        \label{fig-avg}
     \end{subfigure}
    \caption{Comparison of RUL prediction results of four machines using the model.}
    \label{fig:main}
\end{figure*}

In contrast, Figure~\ref{fig-avg} shows that while the federated learning models (Algorithm~\ref{algorithm-fl-transformer}) generate reasonably good average RUL values across multiple machines, they may not capture the intricate details as effectively as single-machine models. However, Figures~\ref{fig-model-001} to \ref{fig-model-004} show a significant bias when models trained on a single machine are applied to other machines, underscoring the challenge of model generalizability in diverse operational environments. This analysis illustrates the trade-off between the granularity and specificity of single-machine models and the broader applicability but potentially lower precision of federated learning approaches, highlighting the need for context-aware modeling strategies in predictive maintenance.

\begin{figure*}[!ht]
     \centering
     \begin{subfigure}[b]{0.47\textwidth}
         \centering
        \includegraphics[width=\linewidth]{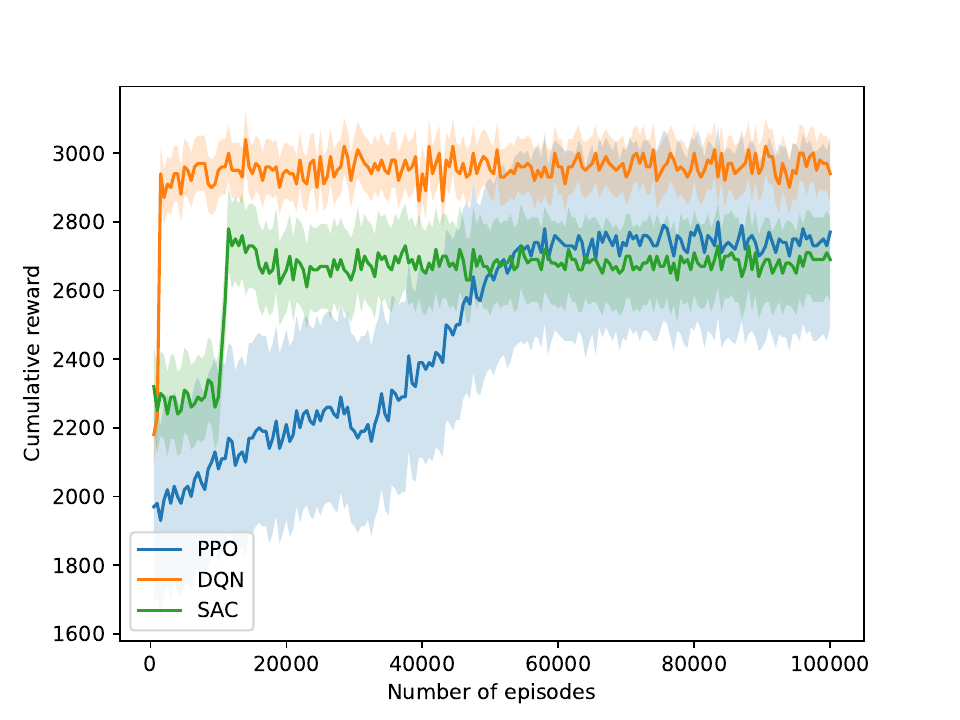}
        \caption{The rewards for $10^5$  episodes during the training of DQN, PPO, and SAC algorithms in our designed environment.}
        \label{fig-drl}
     \end{subfigure}
     \hfill
     \begin{subfigure}[b]{0.47\textwidth}
         \centering
        \includegraphics[width=\linewidth]{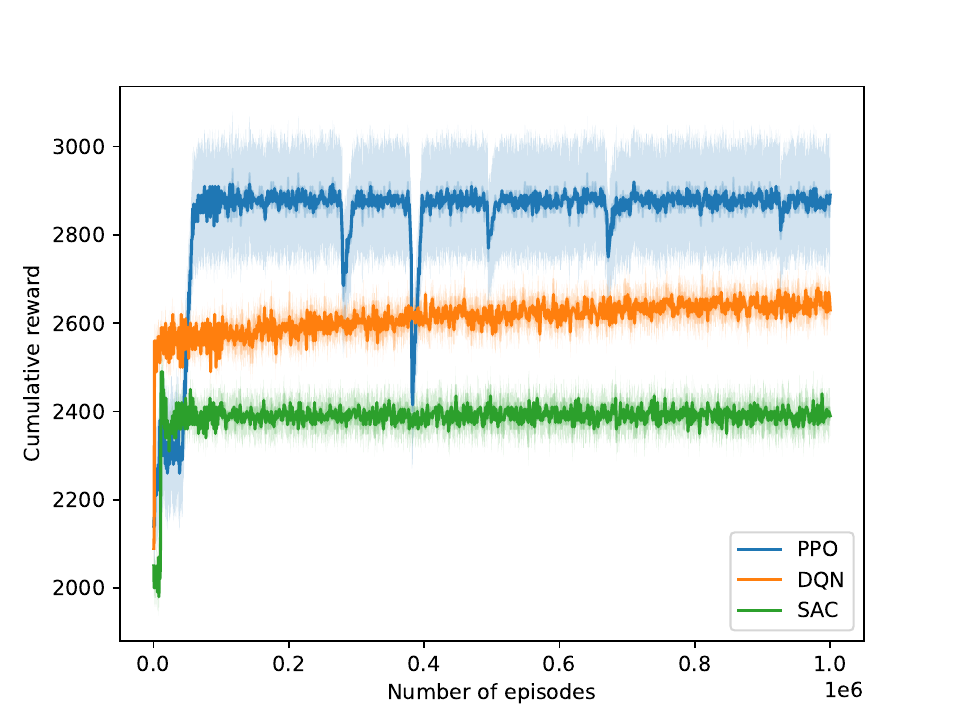}
        \caption{The rewards for $10^6$  episodes during the training of DQN, PPO, and SAC algorithms in our designed environment. }
        \label{fig-drl-6}
     \end{subfigure}
    \caption{Results of Using Deep Reinforcement Learning for Optimizing Maintenance Strategies.}
    \label{fig:main}
\end{figure*}

\begin{figure*}[!ht]
     \centering
     \begin{subfigure}[b]{0.47\textwidth}
         \centering
        \includegraphics[width=\linewidth]{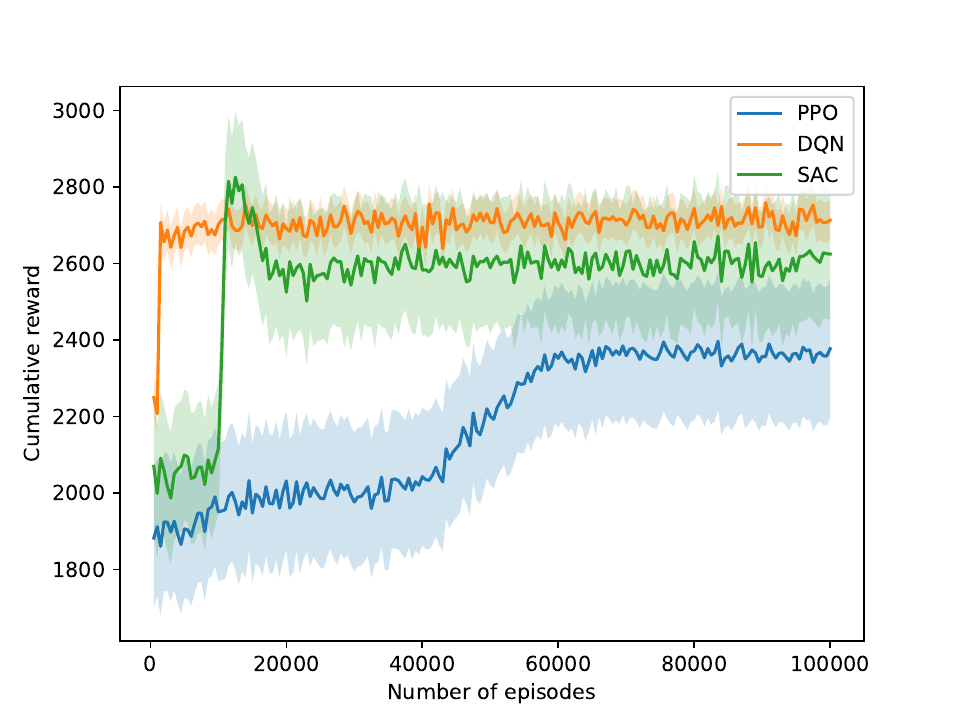}
        \caption{The rewards for $10^5$  episodes during the training of DQN, PPO, and SAC algorithms with human feedback in our designed environment. }
        \label{fig-hf-1}
     \end{subfigure}
     \hfill
     \begin{subfigure}[b]{0.47\textwidth}
         \centering
        \includegraphics[width=\linewidth]{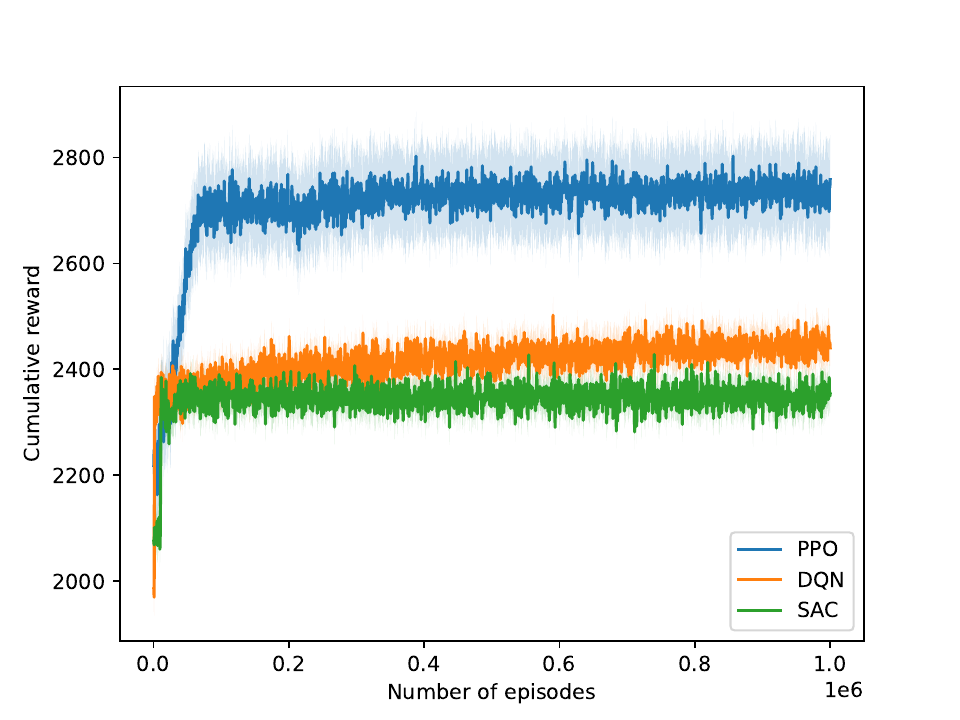}
        \caption{The rewards for $10^6$  episodes during the training of DQN, PPO, and SAC algorithms with human feedback in our designed environment.}
        \label{fig-hf-2}
     \end{subfigure}
    \caption{Results for Using Reinforcement Learning with Human Feedback for Optimizing Maintenance Strategies.}
    \label{fig:main}
\end{figure*}

We compare the performance of DQN, PPO, and SAC in the framework of TranDRL (Algorithm~\ref{algorithm-generalized-RL}) in our designed maintenance task in Figures~\ref{fig-drl} and~\ref{fig-drl-6}. The evaluation metric is the total reward obtained by the agent during a fixed number of episodes. The experimental results show that PPO outperforms both DQN and SAC, achieving a higher total reward on the task. The average total reward obtained by PPO is $55,000$, while DQN and SAC achieve an average total reward of $50,000$ and $45,000$, respectively. However, DQN converges faster than PPO and SAC. In general, PPO and SAC are considered to be more sample-efficient than DQN, which means that they can achieve comparable performance with fewer samples. This is because PPO and SAC use more efficient updates that make better use of the available data. In contrast, DQN can be more sensitive to the choice of hyperparameters and the amount of data used during training, which can affect its convergence speed.

However, the performance of these algorithms can vary depending on the specific task and the quality of the implementation. In some cases, DQN may converge faster than PPO or SAC, especially if the action space is discrete and the observations are low-dimensional. On the other hand, if the action space is continuous or the observations are high-dimensional, PPO and SAC may be more suitable and converge faster. Overall, these experimental results demonstrate the importance of selecting the appropriate DRL algorithm and hyperparameters for a specific task to achieve optimal performance.

Figure~\ref{fig-hf-1} and Figure~\ref{fig-hf-2} reflect the changes after adding on the human feedback in the reward function (Algorithm~\ref{algorithm-rlhf-revised}). The results show that the speed of convergence does not improve too much, but the process becomes more stable and variances minimized. This is affected by the weight of human feedback in the reward function as well as the complexity of the task. Since our task converges very fast even without human feedback, there is not much room for improvement. In the future, we will conduct experiments on more complex real-world scenarios to confirm the effectiveness of the RLHF algorithm.

\section{Conclusion and Future Work}
\label{conclusion}
In this paper, we present a prescriptive maintenance framework (TranDRL) that integrates real-time RUL estimations with DRL algorithms for dynamic action recommendations. The framework starts by employing the Transformer network to estimate the RUL distribution of machines (e.g., engines), offering a robust and adaptive data-driven approach to understanding machine conditions in real-time. As the machine operates, our model continues to dynamically update RUL estimates, providing an ever-evolving status of the equipment's condition. We rigorously compare the performance of several state-of-the-art time-series forecasting models, including NBeats and DeepAR, with our proposed method to validate the accuracy of our RUL estimates. For the framework of DRL-based maintenance action recommendations, we demonstrate the efficiency of SAC, DQN, and PPO with human feedback. Each of these algorithms has its strengths and weaknesses, but they are all adapted to working seamlessly with the dynamically updated RUL estimates.

The limitations of our framework include its inability to process image-based data and the lack of metrics to evaluate the accuracy of human experts' contributions. These shortcomings may compromise the performance of the framework. Future work will focus on refining the framework, such as introducing new neural network designs, to address these limitations.

\bibliographystyle{IEEEtran}
\bibliography{references}

\begin{thebibliography}{10}
\providecommand{\url}[1]{#1}
\csname url@samestyle\endcsname
\providecommand{\newblock}{\relax}
\providecommand{\bibinfo}[2]{#2}
\providecommand{\BIBentrySTDinterwordspacing}{\spaceskip=0pt\relax}
\providecommand{\BIBentryALTinterwordstretchfactor}{4}
\providecommand{\BIBentryALTinterwordspacing}{\spaceskip=\fontdimen2\font plus
\BIBentryALTinterwordstretchfactor\fontdimen3\font minus \fontdimen4\font\relax}
\providecommand{\BIBforeignlanguage}[2]{{%
\expandafter\ifx\csname l@#1\endcsname\relax
\typeout{** WARNING: IEEEtran.bst: No hyphenation pattern has been}%
\typeout{** loaded for the language `#1'. Using the pattern for}%
\typeout{** the default language instead.}%
\else
\language=\csname l@#1\endcsname
\fi
#2}}
\providecommand{\BIBdecl}{\relax}
\BIBdecl

\bibitem{doi:10.1080/00207543.2018.1444806}
E.~L.~X. Li~Da~Xu and L.~Li, ``Industry 4.0: state of the art and future trends,'' \emph{International Journal of Production Research}, vol.~56, no.~8, pp. 2941--2962, 2018.

\bibitem{dangut2022application}
M.~D. Dangut, I.~K. Jennions, S.~King, and Z.~Skaf, ``Application of deep reinforcement learning for extremely rare failure prediction in aircraft maintenance,'' \emph{Mechanical Systems and Signal Processing}, vol. 171, p. 108873, 2022.

\bibitem{5413482}
D.~A. Tobon-Mejia, K.~Medjaher, and N.~Zerhouni, ``The {ISO} 13381-1 standard's failure prognostics process through an example,'' in \emph{2010 Prognostics and System Health Management Conference}, 2010, pp. 1--12.

\bibitem{aitken2021understanding}
K.~Aitken, V.~Ramasesh, Y.~Cao, and N.~Maheswaranathan, ``Understanding how encoder-decoder architectures attend,'' \emph{Advances in Neural Information Processing Systems}, vol.~34, pp. 22\,184--22\,195, 2021.

\bibitem{ong2021deep}
K.~S.~H. Ong, W.~Wang, D.~Niyato, and T.~Friedrichs, ``Deep-reinforcement-learning-based predictive maintenance model for effective resource management in industrial {IoT},'' \emph{IEEE Internet of Things Journal}, vol.~9, no.~7, pp. 5173--5188, 2021.

\bibitem{NIPS2017_faafda66}
N.~Ding and R.~Soricut, ``Cold-start reinforcement learning with softmax policy gradient,'' in \emph{Advances in Neural Information Processing Systems}, I.~Guyon, U.~V. Luxburg, S.~Bengio, H.~Wallach, R.~Fergus, S.~Vishwanathan, and R.~Garnett, Eds., vol.~30.\hskip 1em plus 0.5em minus 0.4em\relax Curran Associates, Inc., 2017.

\bibitem{zonta2020predictive}
T.~Zonta, C.~A. Da~Costa, R.~da~Rosa~Righi, M.~J. de~Lima, E.~S. da~Trindade, and G.~P. Li, ``Predictive maintenance in the industry 4.0: A systematic literature review,'' \emph{Computers \& Industrial Engineering}, vol. 150, p. 106889, 2020.

\bibitem{zhou2014remaining}
Q.~Zhou, J.~Son, S.~Zhou, X.~Mao, and M.~Salman, ``Remaining useful life prediction of individual units subject to hard failure,'' \emph{IIE Transactions}, vol.~46, no.~10, pp. 1017--1030, 2014.

\bibitem{man2018prediction}
J.~Man and Q.~Zhou, ``Prediction of hard failures with stochastic degradation signals using wiener process and proportional hazards model,'' \emph{Computers \& Industrial Engineering}, vol. 125, pp. 480--489, 2018.

\bibitem{jia2016deep}
F.~Jia, Y.~Lei, J.~Lin, X.~Zhou, and N.~Lu, ``Deep neural networks: A promising tool for fault characteristic mining and intelligent diagnosis of rotating machinery with massive data,'' \emph{Mechanical systems and signal processing}, vol.~72, pp. 303--315, 2016.

\bibitem{zhao2019deep}
R.~Zhao, R.~Yan, Z.~Chen, K.~Mao, P.~Wang, and R.~X. Gao, ``Deep learning and its applications to machine health monitoring,'' \emph{Mechanical Systems and Signal Processing}, vol. 115, pp. 213--237, 2019.

\bibitem{vaswani2017attention}
A.~Vaswani, N.~Shazeer, N.~Parmar, J.~Uszkoreit, L.~Jones, A.~N. Gomez, {\L}.~Kaiser, and I.~Polosukhin, ``Attention is all you need,'' \emph{Advances in neural information processing systems}, vol.~30, 2017.

\bibitem{zhou2021informer}
H.~Zhou, S.~Zhang, J.~Peng, S.~Zhang, J.~Li, H.~Xiong, and W.~Zhang, ``Informer: Beyond efficient transformer for long sequence time-series forecasting,'' in \emph{Proceedings of the AAAI conference on artificial intelligence}, vol.~35, no.~12, 2021, pp. 11\,106--11\,115.

\bibitem{huang2020deep}
J.~Huang, Q.~Chang, and J.~Arinez, ``Deep reinforcement learning based preventive maintenance policy for serial production lines,'' \emph{Expert Systems with Applications}, vol. 160, p. 113701, 2020.

\bibitem{gordon2022data}
C.~A. Gordon and E.~N. Pistikopoulos, ``Data-driven prescriptive maintenance toward fault-tolerant multiparametric control,'' \emph{AIChE Journal}, vol.~68, no.~6, p. e17489, 2022.

\bibitem{ansari2019prima}
F.~Ansari, R.~Glawar, and T.~Nemeth, ``{PriMa}: a prescriptive maintenance model for cyber-physical production systems,'' \emph{International Journal of Computer Integrated Manufacturing}, vol.~32, no. 4-5, pp. 482--503, 2019.

\bibitem{nemeth2018prima}
T.~Nemeth, F.~Ansari, W.~Sihn, B.~Haslhofer, and A.~Schindler, ``{PriMa-X}: A reference model for realizing prescriptive maintenance and assessing its maturity enhanced by machine learning,'' \emph{Procedia CIRP}, vol.~72, pp. 1039--1044, 2018.

\bibitem{hakim1998developing}
C.~Hakim, ``Developing a sociology for the twenty-first century: Preference theory,'' \emph{The British journal of sociology}, vol.~49, no.~1, pp. 137--143, 1998.

\bibitem{christiano2017deep}
P.~F. Christiano, J.~Leike, T.~Brown, M.~Martic, S.~Legg, and D.~Amodei, ``Deep reinforcement learning from human preferences,'' \emph{Advances in neural information processing systems}, vol.~30, 2017.

\bibitem{ibarz2018reward}
B.~Ibarz, J.~Leike, T.~Pohlen, G.~Irving, S.~Legg, and D.~Amodei, ``Reward learning from human preferences and demonstrations in atari,'' \emph{Advances in neural information processing systems}, vol.~31, 2018.

\bibitem{lee2021pebble}
K.~Lee, L.~Smith, and P.~Abbeel, ``Pebble: Feedback-efficient interactive reinforcement learning via relabeling experience and unsupervised pre-training,'' \emph{arXiv preprint arXiv:2106.05091}, 2021.

\bibitem{park2022surf}
J.~Park, Y.~Seo, J.~Shin, H.~Lee, P.~Abbeel, and K.~Lee, ``{SURF}: Semi-supervised reward learning with data augmentation for feedback-efficient preference-based reinforcement learning,'' \emph{arXiv preprint arXiv:2203.10050}, 2022.

\bibitem{early2022non}
J.~Early, T.~Bewley, C.~Evers, and S.~Ramchurn, ``Non-markovian reward modelling from trajectory labels via interpretable multiple instance learning,'' \emph{Advances in Neural Information Processing Systems}, vol.~35, pp. 27\,652--27\,663, 2022.

\bibitem{patti2008shape}
A.~L. Patti, K.~Watson, and J.~H. Blackstone~Jr, ``The shape of protective capacity in unbalanced production systems with unplanned machine downtime,'' \emph{Production Planning and Control}, vol.~19, no.~5, pp. 486--494, 2008.

\bibitem{wen2022transformers}
Q.~Wen, T.~Zhou, C.~Zhang, W.~Chen, Z.~Ma, J.~Yan, and L.~Sun, ``Transformers in time series: {A} survey,'' \emph{arXiv preprint arXiv:2202.07125}, 2022.

\bibitem{saxena2008damage}
A.~Saxena, K.~Goebel, D.~Simon, and N.~Eklund, ``Damage propagation modeling for aircraft engine run-to-failure simulation,'' in \emph{2008 international conference on prognostics and health management}.\hskip 1em plus 0.5em minus 0.4em\relax IEEE, 2008, pp. 1--9.

\bibitem{Heimes2008RecurrentNN}
F.~Heimes, ``Recurrent neural networks for remaining useful life estimation,'' \emph{2008 International Conference on Prognostics and Health Management}, pp. 1--6, 2008.

\bibitem{oreshkin2019n}
B.~N. Oreshkin, D.~Carpov, N.~Chapados, and Y.~Bengio, ``{N-BEATS}: Neural basis expansion analysis for interpretable time series forecasting,'' \emph{arXiv preprint arXiv:1905.10437}, 2019.

\bibitem{salinas2020deepar}
D.~Salinas, V.~Flunkert, J.~Gasthaus, and T.~Januschowski, ``{DeepAR}: Probabilistic forecasting with autoregressive recurrent networks,'' \emph{International Journal of Forecasting}, vol.~36, no.~3, pp. 1181--1191, 2020.

\bibitem{fan2020theoretical}
J.~Fan, Z.~Wang, Y.~Xie, and Z.~Yang, ``A theoretical analysis of deep q-learning,'' in \emph{Learning for dynamics and control}.\hskip 1em plus 0.5em minus 0.4em\relax PMLR, 2020, pp. 486--489.

\bibitem{schulman2017proximal}
J.~Schulman, F.~Wolski, P.~Dhariwal, A.~Radford, and O.~Klimov, ``Proximal policy optimization algorithms,'' \emph{arXiv preprint arXiv:1707.06347}, 2017.

\bibitem{haarnoja2018soft}
T.~Haarnoja, A.~Zhou, P.~Abbeel, and S.~Levine, ``Soft actor-critic: Off-policy maximum entropy deep reinforcement learning with a stochastic actor,'' in \emph{International conference on machine learning}.\hskip 1em plus 0.5em minus 0.4em\relax PMLR, 2018, pp. 1861--1870.

\end{thebibliography}

\end{document}